\documentclass[10pt,twocolumn,letterpaper]{article}

\usepackage{iccv}
\usepackage{times}
\usepackage{epsfig}
\usepackage{graphicx}
\usepackage{amsmath}
\usepackage{amssymb}
\usepackage{multirow}
\usepackage[pagebackref=true,breaklinks=true,colorlinks,bookmarks=false]{hyperref}

\iccvfinalcopy % *** Uncomment this line for the final submission

\ificcvfinal\fi

\title{Zero-Shot  Contrastive Loss for Text-Guided Diffusion  Image Style Transfer}

\author{Serin Yang{} ,   Hyunmin Hwang,   Jong Chul Ye\\
Kim Jaechul Graduate School of AI\\
Korea Advanced Institute of Science and Technology (KAIST)\\
{\tt\small \{yangsr, hyunmin\_hwang, jong.ye\}@kaist.ac.kr}
}
\begin{document}

\begin{figure*}
\twocolumn[{
\renewcommand\twocolumn[1][]{#1}
\maketitle
    \centerline{\includegraphics[width=\textwidth]{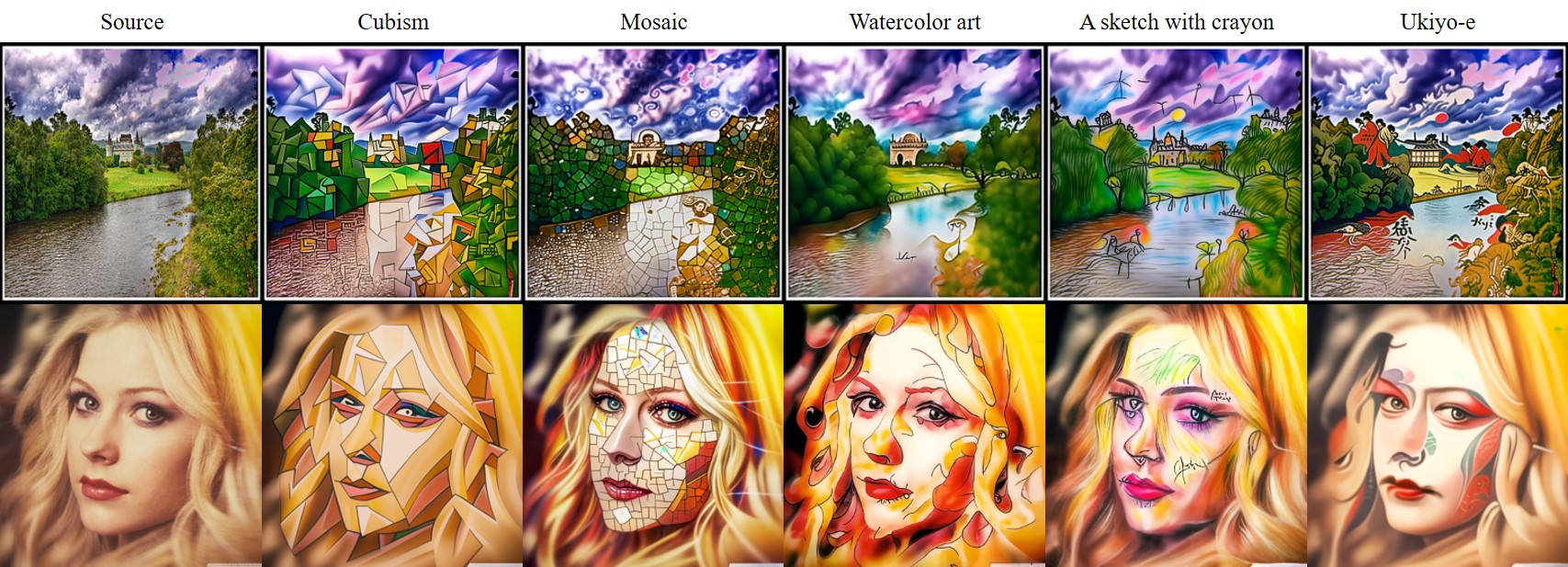}}
    \caption{Our style transfer method produces impressive results when applied to a range of artistic styles. The method preserves the underlying structure of the source images while transforming them into the desired styles.}
    \label{fig_result_artstyle}
    \vspace{0.3cm}
}]
\end{figure*}

\ificcvfinal\thispagestyle{empty}\fi

%%%%%%%%% ABSTRACT
\begin{abstract}
%\vspace{-0.5cm}
Diffusion models have shown great promise in text-guided image style transfer, but there is a trade-off between style transformation and content preservation due to their stochastic nature. Existing methods require computationally expensive fine-tuning of diffusion models or additional neural network. To address this, here we propose a zero-shot contrastive   loss for diffusion models that doesn't require additional fine-tuning or auxiliary networks. By leveraging patch-wise contrastive loss between generated samples and original image embeddings in the pre-trained diffusion model, our method can generate images with the same semantic content as the source image in a zero-shot manner. Our approach outperforms existing methods while preserving content and requiring no additional training, not only for image style transfer but also for image-to-image translation and manipulation. Our experimental results validate the effectiveness of our proposed method. Code is available at \href{https://github.com/ZeConloss/ZeCon}{https://github.com/ZeConloss/ZeCon} 
\end{abstract}

%%%%%%%%% BODY TEXT
\section{Introduction}
Style transfer is the task that converts the style of a given image into another style while preserving its content. Over the past few years, GAN-based methods such as pix2pix~\cite{isola2017image}, cycleGAN~\cite{zhu2017unpaired}, and contrastive unpaired image-to-image translation (CUT) have been developed~\cite{park2020contrastive}. 
Recently, joint use of a pretrained image generator and image-text encoder enabled text-guided image editing which requires little or no training of the networks~\cite{radford2021learning,crowson2022vqgan,Patashnik_2021_ICCV,gal2022stylegan,kwon2022clipstyler}.

% Diffusion model의 이슈. 좋은 성능. 다양한 task에 적용.
Inspired by the success of diffusion models for image generation~\cite{ho2020denoising,song2020denoising}, image editing~\cite{liu2021more}, in-painting~\cite{avrahami2022blended}, super-resolution~\cite{chung2022come}, etc., 
many researchers have recently investigated the application of the diffusion models for image-to-image style transfer~\cite{saharia2022palette, su2022dual}.
For example,~\cite{saharia2022palette,saharia2021image} proposed
conditional diffusion models  that require paired dataset for image-to-image style transfer.
One of the limitations of these approaches is that 
the diffusion models need to be trained with paired data set with matched source and target styles.
As collecting matched source and target domain data is impractical,
many recent researchers have focused on unconditional diffusion models.
Unconditional diffusion models have limitations in maintaining content due to the stochastic nature of the reverse sampling procedure that doesn't explicitly impose content consistency. As a result, content and styles can change simultaneously, creating challenges for maintaining content. 

To tackle this problem, the dual diffusion implicit bridge (DDIB)~\cite{su2022dual} exploits two score functions that have been independently trained on two different domains. Although DDIB can translate one image into another without any external condition, it also requires training of two diffusion models for each domain which involves additional training time and a large amount of dataset.  On the other hand,
 DiffusionCLIP~\cite{kim2022diffusionclip} leverages the pretrained diffusion models and CLIP encoder to enable
 text-driven image style transfer without additional large training data set. Unfortunately,
DiffusionCLIP still requires additional fine-tuning of the model for the desired style. 
Furthermore, DiffuseIT \cite{kwon2022diffusion} uses disentangled style and content representation inspired by the slicing Vision Transformer \cite{tumanyan2022splicing}.  Although DiffuseIT has shown its superiority in preserving content, it still suffers from the trade-off between transforming the texture of images and maintaining the content. Also, an additional network is required for computing content losses in DiffuseIT.

To address this problem, here we propose a simple yet effective  Zero-shot Contrastive (ZeCon)  loss  for diffusion models to transfer the style of a given image while preserving its semantic content in a zero-shot manner. Our approach is based on the observation that a pre-trained diffusion model already contains spatial information in its embedding that can be used to maintain content through patch-wise contrastive loss between the input image and generated images. Unlike DiffusionCLIP, {our method doesn't require additional training. In other words, we could effectively preserve the content in a zero-shot manner by leveraging the patch-wise contrastive loss. Furthermore, unlike DiffuseIT, our method achieves more accurate texture modification while preserving the content.}

To demonstrate the effectiveness of our proposed method, we show a text-driven style transfer using CLIP~\cite{radford2021learning}. However, our method can be extended for general guidance beyond text inputs. Furthermore, we demonstrate that our method can be applied to text-driven image-to-image translation and image manipulation tasks, illustrating its wide applicability.

\section{Related Works}
\label{related_works}

\paragraph{Image style transfer}

Neural style transfer~\cite{gatys2016image} iteratively optimizes the content image to match the style image, which is time-consuming. Alternatively, adaptive instance normalization (AdaIN)~\cite{huang2017arbitrary} transfers the style of a source image to a target image by matching their feature statistics.

In contrast, pix2pix~\cite{isola2017image}, CycleGAN~\cite{zhu2017unpaired}, and CUT~\cite{park2020contrastive} use different mechanisms for content preservation. CycleGAN's cycle consistency for content preservation is often too restrictive, while CUT maximizes mutual information between content and stylized images in a patch-based feature space. This maintains structural information while changing appearance.

%Neural style transfer ~\cite{gatys2016image} is the first approach to change the style texture of the content image into a style image by iterative optimization process. However, this iterative process takes a significant amount of time. 
%Alternatively, the adaptive instance normalization (AdaIN) by~\cite{huang2017arbitrary} converts the means and variances of the features of the source image to those of the target image, which enables arbitrary style transfer. 
%
%
%On the other hand, pix2pix~\cite{isola2017image}, CycleGAN~\cite{zhu2017unpaired}  and CUT~\cite{park2020contrastive} rely on different mechanisms for content preservation.
%Specifically, in CycleGAN~\cite{zhu2017unpaired}, 
%the cycle consistency assumes a bijective relationship between two domains for content preservation, whose constraint is often restrictive in some applications.
%In order to overcome this restriction, CUT~\cite{park2020contrastive} was proposed to maximize the mutual information between the content input and stylized output images in a patch-based manner on the feature space. This leads to preservation of the structure between the two images while changing appearance.
%
% text-guided image manipulation with CLIP
CLIP model~\cite{radford2021learning} has been shown to have semantic representative power resulting from a large-scale dataset of 400 million image and text pairs, which allows for text-driven image manipulation. StyleCLIP~\cite{Patashnik_2021_ICCV} uses CLIP and pretrained StyleGAN~\cite{karras2020analyzing} to optimize the latent vector of the content input given a text prompt, but its image modification is limited to the domain of the pretrained generator. StyleGAN-NADA~\cite{gal2022stylegan} proposes an out-of-domain image manipulation method that shifts the generative model to new domains. VQGAN-CLIP~\cite{crowson2022vqgan} demonstrates that VQGAN~\cite{esser2021taming} can also be used as a pretrained generative model to generate or edit high-quality images without training. CLIPstyler~\cite{kwon2022clipstyler} proposes a CNN encoder-decoder model that learns both content and style properties through patch-wise CLIP loss, allowing for image generation and manipulation beyond the domains of pretrained generators.

%With the advent of CLIP model~\cite{radford2021learning}, it has been shown that text-guided image synthesis can be accomplished without collecting style images. CLIP has semantic representative power which results from a large scale dataset consisting of 400 million image and text pairs. This enables text-driven image manipulation. 
%StyleCLIP~\cite{Patashnik_2021_ICCV} was proposed to optimize the latent vector of the content input given a text prompt by using CLIP and pretrained StyleGAN~\cite{karras2020analyzing}. However, image modification using StyleCLIP is limited to the domain of the pretrained generator. 
%In order to solve this issue, StyleGAN-NADA~\cite{gal2022stylegan} presented an out-of-domain image manipulation method that shifts the generative model to new domains. 
%VQGAN-CLIP~\cite{crowson2022vqgan} has shown that VQGAN~\cite{esser2021taming} can also be used as a pretrained generative model to generate or edit high quality images without training.
%In order not to be restricted to the domains of the pretrained generators, CLIPstyler~\cite{kwon2022clipstyler} proposed a CNN encoder-decoder model that learns both content and style properties through patch-wise CLIP loss. 

\begin{figure*}[!t]
	\centerline{\includegraphics[width=1.8\columnwidth]{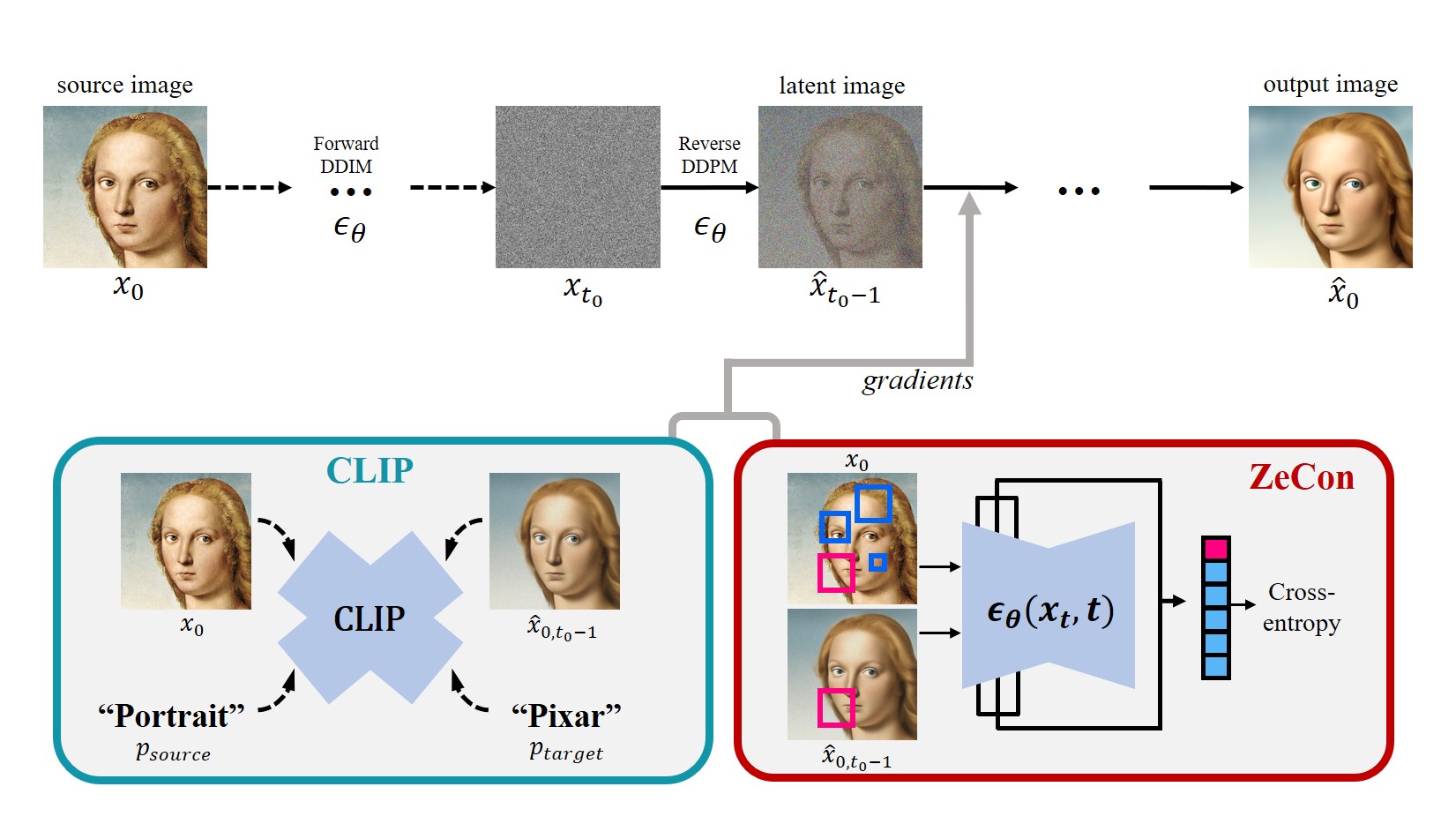}}
	\caption{Our proposed method. To guide the diffusion model in our proposed method, we calculate the ZeCon loss using a noise estimator $\epsilon_{\theta}(\cdot)$ and the CLIP loss using the CLIP model. These losses allow us to add gradients to the denoised image at each time step. }
	\label{fig_scheme2}
\end{figure*}

\paragraph{Diffusion models for image style transfer}
% diffusion models 전반에 대한 간략한 소개 및 설명

Diffusion models have become popular due to their impressive ability to generate high-quality images ~\cite{ho2020denoising, song2020denoising}. Diffusion models have found application in various computer vision areas, including super-resolution~\cite{rombach2022high}, segmentation~\cite{baranchuk2022labelefficient}, image editing~\cite{avrahami2022blended}, medical image processing~\cite{kim2021diffusemorph}, and video generation~\cite{ho2022video}.

This generative model works by progressively adding Gaussian noise through a Markov chain forward process. Then, a trained noise estimation model is used to generate clean samples from the latent noise through an iterative denoising process.
Specifically, % in diffusion models, given a source data distribution $q(x_0)$, latent variable $x_t$ is computed by forward diffusion process. 
DDPM~\cite{ho2020denoising} directly samples $x_t$ from $x_0$ by adding Gaussian noise with $\beta_t \in (0, 1)$ at time $t \in [1,...,T]$,
\begin{equation} \label{eq_ddpm_forward}
x_t = \sqrt{\overline{\alpha}_t}x_0 + \sqrt{1-\overline{\alpha}_t}\epsilon
\end{equation}
where $\epsilon \sim \mathcal{N}(0,I)$, $\alpha_t = 1 - \beta_t$, and $\overline{\alpha}_t = \prod_{i=0}^{t}\alpha_i$. The reverse sampling process to generate a clean image is then given by:
\begin{equation} \label{eq_ddpm_reverse}
x_{t-1} 
= \frac{1}{\sqrt{1-\beta_t}}
\left(x_t - \frac{\beta_t}{\sqrt{1-\overline{\alpha}_t}}\epsilon_{\theta}(x_t,t)\right)
+\sigma_{t}\epsilon.
\end{equation}
where the neural network $\epsilon_{\theta}(x_t,t)$ is used to estimate the noise component, which can be viewed as a score function up to a scaling factor.

\begin{figure}[!t]
	\centerline{\includegraphics[width=\columnwidth]{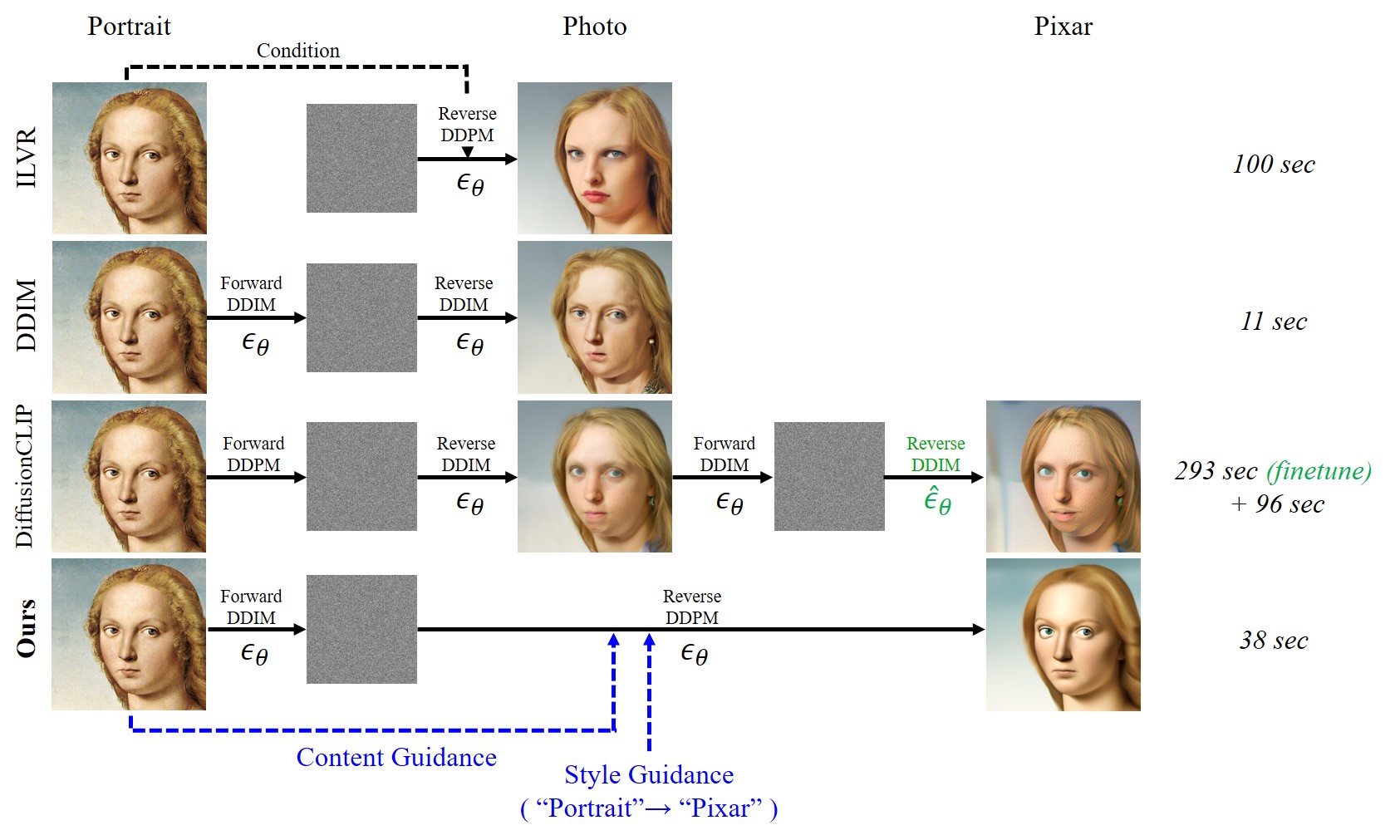}}
	\caption{An illustration on sampling schemes of four diffusion models for style transfer. }
	\label{fig_schemes}
\end{figure}

While noise $\epsilon$ can help to achieve sample diversity in DDPM, it may also lead to a loss of content in the context of style transfer. The repeated application of stochastic operations can result in images with completely different content, even if the intermediate latent space is the same for each image.
The content can be preserved with DDIM~\cite{song2020denoising} whose sampling process is:
\begin{equation} \label{eq_ddim_reverse}
\begin{aligned}
x_{t-1} 
& = \sqrt{\overline{\alpha}_{t-1}}\hat x_{0,t}(x_t) \\
& \indent + \sqrt{1-\overline{\alpha}_{t-1}-\sigma_t^2}\epsilon_{\theta}(x_t,t) 
+ {\sigma_t^2}\epsilon
\end{aligned}
\end{equation}
where $\sigma_t$ is the variance of noise which controls how stochastic the sampling process is, and $\hat x_{0,t}$ is a denoised image given by:
\begin{equation} \label{eq_ddim_clean}
\hat x_{0,t}(x_t) := \frac{x_t - \sqrt{1-\overline{\alpha}_t}\epsilon_{\theta}(x_t,t)}{\sqrt{\overline{\alpha}_t}}.
\end{equation}
If we set $\sigma_t = 0$ in  \eqref{eq_ddim_reverse}, the noise term $\epsilon$ is eliminated, which allows us to preserve the content successfully. However, in this case, the sampling process becomes deterministic, which results in preserving the style as well. This is not desirable for style transfer, as illustrated in Figure~\ref{fig_schemes}.

To preserve semantics in style transfer, Palette~\cite{saharia2022palette} used conditional diffusion models that require paired datasets for training $\epsilon_{\theta}(x_t,t)$. On the other hand, ILVR~\cite{Choi_2021_ICCV} attempted to generate diverse samples conditioned on the input image using unconditional models, but the stochasticity introduced by $\epsilon$ still posed a challenge, as shown in Figure~\ref{fig_schemes}. Unconditional models were also employed in DDIB~\cite{su2022dual}, which used two independently trained score functions for different domains, and in DiffusionCLIP~\cite{kim2022diffusionclip}, which fine-tuned a pre-trained diffusion model using identity and style losses. However, both methods require training or fine-tuning diffusion models for each style domain. Furthermore, 
DiffuseIT \cite{kwon2022diffusion} necessitates an auxiliary network for computing a content loss.

\section{Main Contributions}

\paragraph{Sampling strategy}
Similar to \eqref{eq_ddim_reverse}, DDPM can be also represented as
\begin{equation} \label{eq_ddpm_reverse2}
\begin{aligned}
x_{t-1} 
& = \sqrt{\overline{\alpha}_{t-1}}\hat x_{0,t}(x_t) \\
& \indent + \sqrt{1-\overline{\alpha}_{t-1}-\sigma_t^2}\epsilon_{\theta}(x_t,t) 
+ {\sigma_t^2}\epsilon
\end{aligned}
\end{equation}
if  $\sigma_t$ is given by
\begin{align}
\sigma_t=\sqrt{(1-\bar\alpha_{t-1})/(1-\bar\alpha_t)}\sqrt{1-\bar\alpha_t/\bar\alpha_{t-1}}
\end{align}
We then define the loss function as follows
\begin{align}
 \ell_{total}(x) =   \ell_{content}(x)  + \ell_{CLIP}(x)
\end{align}
where $\ell_{content}$ and $\ell_{CLIP}$ denotes the content and style loss, respectively.
Then,  the denoised image estimate $\hat x_{0,t}(x_t)$ is supplemented
with the gradient of the loss function:
\begin{equation} \label{eq_updated_x}
\hat x_{0,t}(x_t) = \hat x_{0,t}(x_t) +\nabla_{x} \ell_{total} (x)|_{x=\hat x_{0,t}(x_t)}
\end{equation}
{Unlike DiffuseIT, which requires a substantial auxiliary network for computing content loss, our approach relies on a simpler but still effective content loss, as detailed below.}

%
%\paragraph{Style guidance for diffusion models}
%
%Guiding gradients in diffusion models is the method proposed in the context of class-conditional image generation~\cite{dhariwal2021diffusion}. 
%Accordingly, even the unconditional diffusion model can generate conditional images using guidance by classifiers or CLIP. 
%
%Specifically,  in  DDPM, the estimated clean image $\hat{x}_{0,t}$ from the latent variable $x_t$~\cite{avrahami2022blended} can be produced from the noise approximation model $\epsilon_{\theta}(x_t,t)$:
%\begin{equation} \label{eq_ddpm_clean}
%\displaystyle \hat{x}_{0,t} = \frac{x_t} {\sqrt{\overline{\alpha}_t}} - \frac{\sqrt{1-\overline{\alpha}_t}\epsilon_{\theta}(x_t, t)}{\sqrt{\overline{\alpha}_t}}.
%\end{equation}
%Then, 
%as proposed in the past work~\cite{avrahami2022blended}, 
%The gradients are then added to the predicted mean at each time step $t$, as  illustrated in Figure \ref{fig_scheme2}.
%More specifically, the gradients for the guidance is calculated as
%\vspace{0.25cm}
%\centerline{
%$\displaystyle \nabla_{\hat{x}_{0,t}} L_{total}(\hat{x}_{0,t},{x_0},p_{target},p_{source})$
%}
%\vspace{0.25cm}
%where $p_{target}$ and $p_{source}$ are text prompts for target and source domain, respectively. 
%
% %Thanks to the guidance method, diffusion models do not require additional training. 
%The guiding method is
%In order to achieve superior performance, we incorporate two types of guidance - style and content guidance. 
%
%
%

\paragraph{Content preservation loss}

In \cite{park2020contrastive}, it was demonstrated that the CUT loss effectively preserves structural information by maximizing the mutual information between input and output patches. Specifically, the original algorithm involves training an encoder to capture spatial information from the input. The resulting encoder features are then used to apply patch-wise contrastive loss, which utilizes the spatial information to preserve the contents.

Recent work by Baranchuk et al. \cite{baranchuk2022labelefficient} has shown that the U-Net noise predictor in the diffusion model contains spatial information. Therefore, a key contribution of this paper is to demonstrate that spatial features required for the CUT loss can be extracted from the diffusion model without additional training, as illustrated in Figure \ref{fig_scheme2}.

Specifically, at each reverse timestep $t \in [t_0,0]$, both the original image $x_0$ and the reverse sampled denoised image $\hat{x}_{0,t}$ are forwarded to the noise estimator $\epsilon_{\theta}(x_t,t)$. The encoder part of the estimator is used to extract feature maps $z_l$ and $\hat{z}_l$ for $x_0$ and $\hat{x}_{0,t}$, respectively.
To apply patch-wise contrastive loss, the pixels of the feature maps are randomly selected and used to calculate cross-entropy loss. Pixels from the same location are considered ``positive" and their mutual information is maximized. Pixels from different locations, considered ``negative", have their mutual information minimized. This process can be expressed mathematically as:
\begin{equation} \label{eq_CUT_loss}
\ell_{ZeCon}(\hat{x}_{0,t},x_0) = \mathbb{E}_{x_0} \left[\sum_{l}\sum_{s}\ell(\hat{z}_\mathit{l}^s,z_\mathit{l}^s,z_\mathit{l}^{S \backslash s}) \right] 
\end{equation}
Here, $\hat{z}_l$ and $z_l$ denote the $l$-th layer features from $\hat{x}_{0,t}$ and $x_0$, respectively. $s$ represents a spatial location in ${1,\dots,S_l}$, where $S_l$ is the number of spatial locations in feature $z_l$. The cross-entropy loss is denoted by $\ell(\cdot)$.
By using the ZeCon loss in \eqref{eq_CUT_loss}, we can maintain semantic consistency between the reverse sampled denoised image $\hat{x}_{0,t}$ and the original image $x_0$, preserving content information. More details can be found in the Supplementary material.

On top of the contrastive loss, we include the feature loss $\ell_{VGG}$, which is the mean-squared error between the VGG feature maps of $\hat{x}_{0,t}$ and $x_0$, and the pixel loss $\ell_{MSE}$, which is the $\ell_2$ norm of the pixel difference between them. 
\begin{equation} \label{eq_guidance_content}
\begin{aligned}
\mathcal{L}_{content} 
& = \ell_{ZeCon}(\hat{x}_{0,t},x_0) \\
&  \indent + \ell_{VGG}(\hat{x}_{0,t},x_0) + \ell_{MSE}(\hat{x}_{0,t},x_0) 
\end{aligned}
\end{equation}
%Therefore, the total loss function for the guidance is formulated as follows: 
%\begin{equation} 
%\begin{aligned}
%\mathcal{L}_{total} & = \mathcal{L}_{CLIP}(\hat{x}_{0,t},x_0,p_{target},p_{source}) \\
%&  \indent + \mathcal{L}_{content}(\hat{x}_{0,t},x_0)
%\end{aligned}
%\end{equation}
The weights for each loss function are hyperparameters which need to be chosen by users. The examples of these weights are given in the Supplementary material.

\paragraph{Style loss}

The CLIP model is trained on extensive language and image dataset which results in its great semantic power~\cite{radford2021learning}. Thanks to this semantic capacity, we can generate images in diverse styles with only text prompts. The CLIP loss for style guidance can be formulated as follows:
\begin{multline} \label{eq_guidance_clip}
%\begin{split}
\ell_{CLIP} = \ell_{global}(\hat{x}_{0,t},p_{target}) \\
+\ell_{dir}(\hat{x}_{0,t},x_0,p_{target},p_{source})
%\end{split}
\end{multline}
Here, the global CLIP loss $\ell_{global}$ calculates the cosine distance in the CLIP embedding space between the generated image $\hat{x}_{0,t}$ and the style prompt $p_{target}$~\cite{Patashnik_2021_ICCV} by
\begin{equation} \label{eq_clip_glob_loss}
\ell_{global}(\hat{x}_{0,t},p_{target}) = D_{CLIP}(\hat{x}_{0,t},p_{target}).
\end{equation}
Since the global loss suffers from mode collapse and corrupted image quality, the directional CLIP loss $\ell_{dir}$ was proposed~\cite{gal2022stylegan}. It aligns the direction in the CLIP embedding space between text and image pairs, which can be formulated as follows:
\begin{equation*}
\ell_{dir}(\hat{x}_{0,t},x_0,p_{target},p_{source}) = 1 - \frac{\Delta I\cdot\Delta T }{\parallel\Delta I\parallel\parallel\Delta T\parallel}
\end{equation*} 
where $p_{source}$ denotes the source text prompt,
and
 $\Delta I=E_{img}(x_0)-E_{img}(\hat{x}_{0,t})$, $\Delta T=E_{txt}(p_{source})-E_{txt}(p_{target})$ for CLIP's image encoder $E_{img}$ and text encoder $E_{txt}$.  
As the patch-based CLIP loss was proposed to enhance the generated images' quality~\cite{kwon2022clipstyler}, we adopt the patch-based scheme in both $\ell_{global}$ and $\ell_{dir}$. 

\begin{figure*}[t]
	\centerline{\includegraphics[width=1.9\columnwidth]{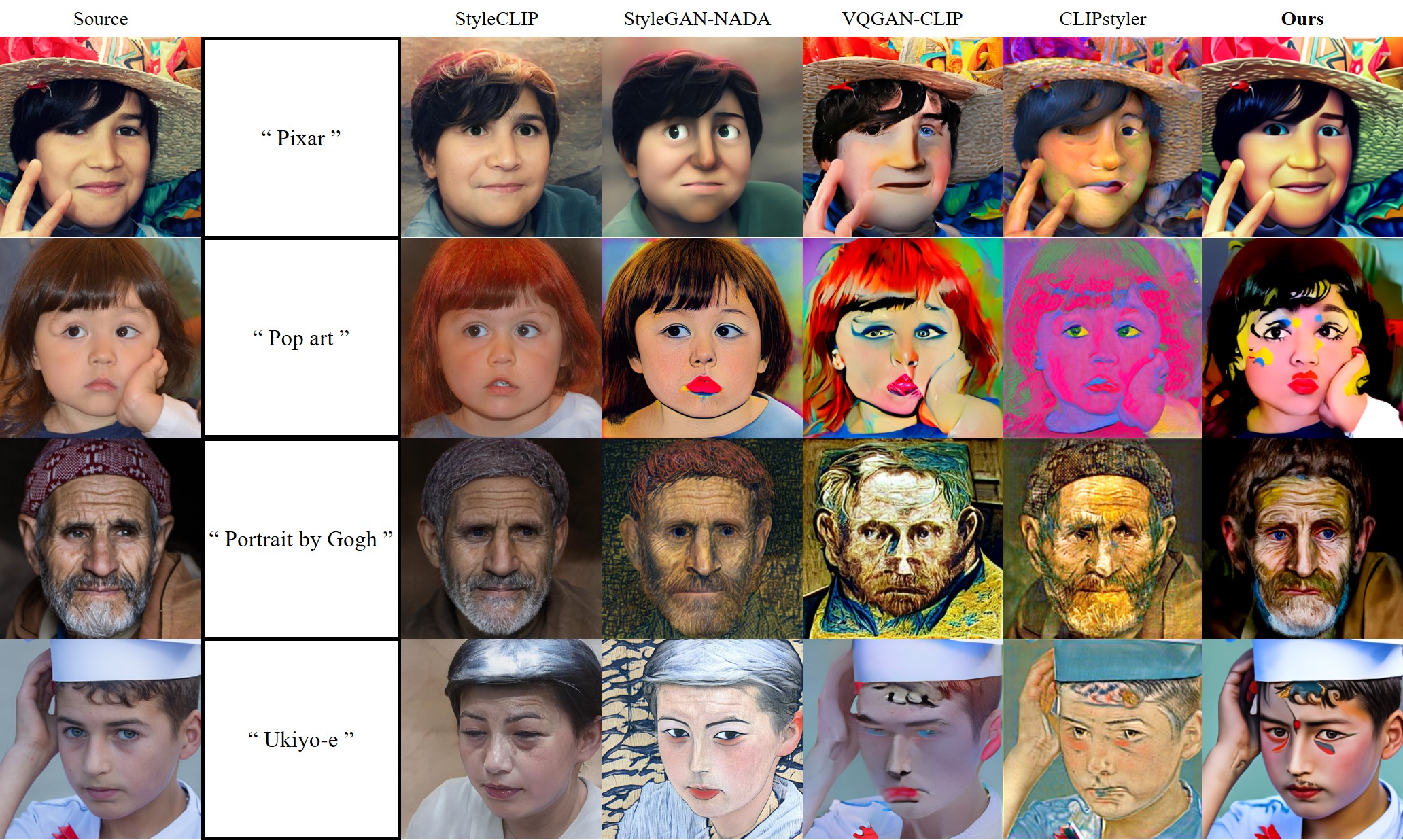}}
	\caption{Comparison against GAN-based style transfer methods. When compared to four GAN-based methods, our approach achieves superior results in terms of style transformation and content preservation.}
	\label{fig_result_Gan}
\end{figure*}

\section{Experimental Results}
\label{experiments}

\subsection{Experimental setting}

\paragraph{Dataset }
The images used as content reference are from FFHQ~\cite{karras2019style}, CelebA-HQ ~\cite{karras2017progressive}, ImageNET~\cite{5206848}, LSUN-Church~\cite{yu2015lsun}, and CycleGAN dataset~\cite{zhu2017unpaired}. They contain images of human faces, objects, scenes, and churches. 
Furthermore, in order to evaluate the performance of our proposed model on the images from unseen domains, we utilize Wikiart dataset~\cite{danielczuk2019segmenting}.
All the images are resized to 256 $\times$ 256 for the diffusion models. For patch-based guidance, we randomly crop 96 patches from a source image and then apply perspective augmentation and affine transformation. More details are illustrated in the Supplementary material.

\paragraph{Diffusion models }
We utilize the pre-trained unconditional diffusion model trained on ImageNET dataset with 256 $\times$ 256 image size~\cite{dhariwal2021diffusion} and the model trained on FFHQ dataset with 256 $\times$ 256 image size~\cite{Choi_2021_ICCV}.

%\textbf{Sampling scheme }
Either DDIM or DDPM method can be applied in our method during the forward and reverse diffusion steps. 
We basically adopt the DDIM strategy as the forward noising process and DDPM method as the reverse sampling.  
When $T$ is the total time step, we respace the step size from $T$ to $T'$. Then with the source image $x_0$, we obtain the latent $x_{t_0}$ from the forward diffusion process, where $t_0 \in [0,T']$. We choose $(T', t_0)$ as $(50, 25)$ as default when $T=1000$. From this latent $x_{t_0}$, the stylized output image is sampled through diffusion processes. 
This approach not only preserves more latent information from the source image, but also enables the image to be effectively converted to a new style. Additionally, by reducing the number of iterations required, inference time can be significantly reduced.

The sampling scheme is illustrated in Figure \ref{fig_scheme2} and \ref{fig_schemes}, and the comparative studies on the choice of $(T', t_0)$ are presented in the Supplementary material.

\subsection{Comparative studies}

Figure \ref{fig_result_artstyle} shows that our method achieves outstanding results across various artistic styles. In addition, we perform comparisons with GAN-based and diffusion-based style transfer methods, respectively.

\begin{figure}[t]
	\centerline{\includegraphics[width=0.9\columnwidth]{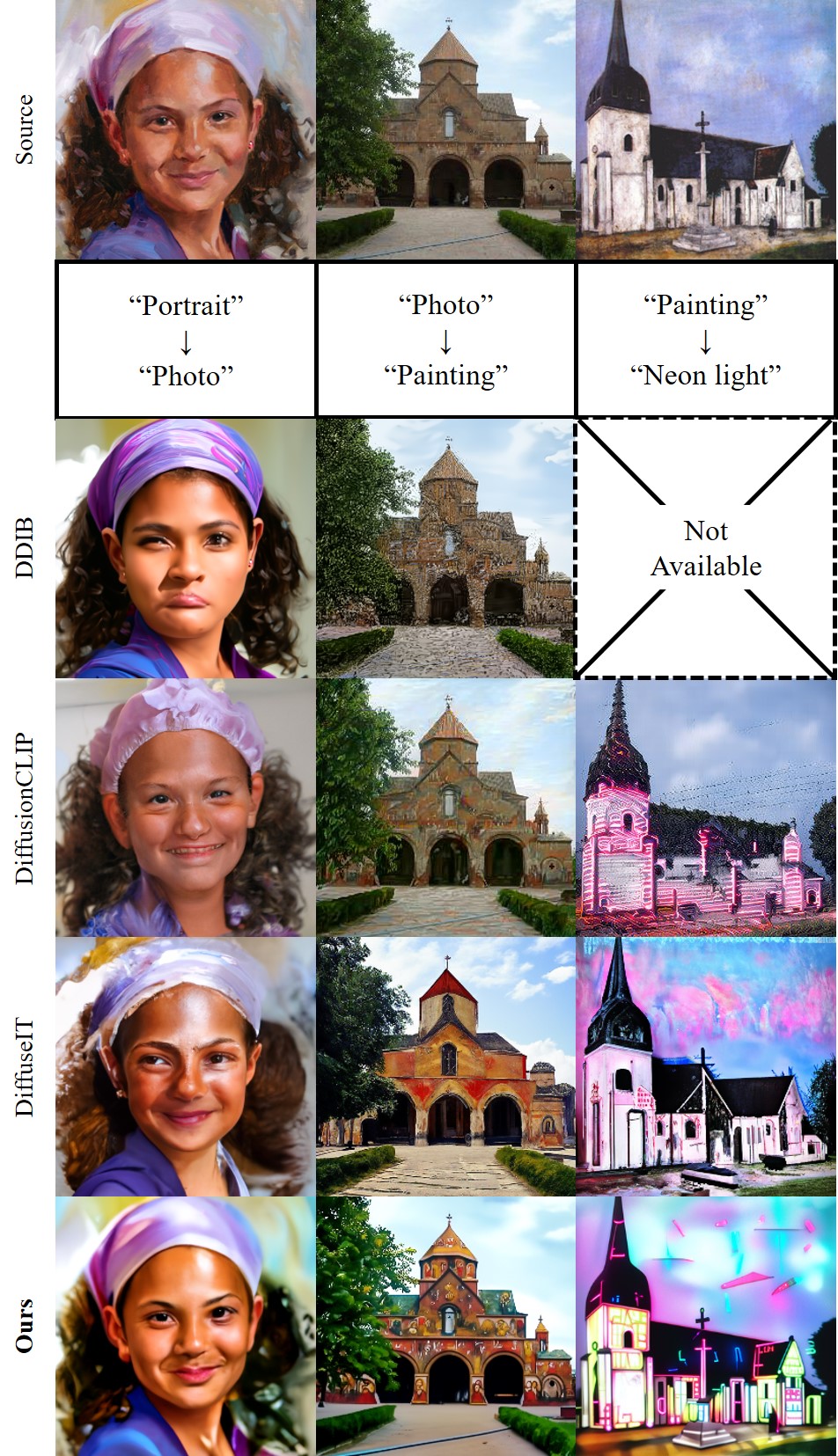}}
	\caption{Comparing three diffusion-based style transfer methods, our proposed approach stands out by allowing for style modulation from unseen domain images, which the other diffusion models cannot achieve.}
	\label{fig_result_diffusion}
\end{figure}

\begin{table}[t]
\centering
\resizebox{\columnwidth}{!}{
\begin{tabular}{c|p{1.5cm}p{1.5cm}|p{2cm}p{2cm}}
    \hline
    \rule{0pt}{1\normalbaselineskip}
    \multirow{2}{*}{Methods}  
    &\multicolumn{2}{|c|}{User study}  
    &\multicolumn{1}{c}{\multirow{2}{*}{CLIP score $\uparrow$}}  
    &\multicolumn{1}{c}{\multirow{2}{*}{Face ID $\downarrow$}} \\ [0.5ex]
    &\multicolumn{1}{|c}{Content $\uparrow$}  
    &\multicolumn{1}{c|}{Style $\uparrow$} & & \\    [0.5ex]
    \hline\hline

    \rule{0pt}{1\normalbaselineskip}
    StyleCLIP   &\hfil \underline{4.10}   &\hfil 1.62   &\hfil 0.0925 &\hfil \bf 0.3750\\
    StyleGAN-NADA   &\hfil 3.42   &\hfil 2.94   &\hfil 0.1222 &\hfil 0.4948\\
    VQGAN-CLIP  &\hfil 1.83   &\hfil 2.92   &\hfil \underline{0.1379} &\hfil 0.7661\\
    CLIPstyler  &\hfil 1.99  &\hfil \underline{2.96}    &\hfil 0.1347 &\hfil 0.6664\\
    Ours    &\hfil \bf 4.61    &\hfil \bf 4.23  &\hfil \bf 0.1479 &\hfil \underline{0.3881} \\ [0.5ex]
    \hline
\end{tabular}
}
\vspace{0.1cm}
\caption{User study and quantitative results for comparison with GAN-based methods for style transfer. The bold text and underline refer to the best and second best results, respectively.}
\label{table_gan}
\end{table}

\paragraph{Comparison with GAN-based models} \label{section_GAN}
For GAN-based models, we compare four state-of-the-art methods - StyleCLIP~\cite{Patashnik_2021_ICCV}, StyleGAN-NADA~\cite{gal2022stylegan}, VQGAN-CLIP~\cite{crowson2022vqgan}, CLIPstyler~\cite{kwon2022clipstyler}. The results of the comparison are illustrated in Figure \ref{fig_result_Gan}. 

Our proposed model clearly outperforms other methods in terms of retaining content. The outputs generated by StyleCLIP and StyleGAN-NADA exhibit distorted results where non-face objects, such as hands or hats, are removed from the output images. While results from VQGAN-CLIP and CLIPstyler show relatively better preservation of facial features, such as eyes and mouth, they still suffer from some loss of detail.

\begin{table}[t]
\centering
\resizebox{\columnwidth}{!}{
    \begin{tabular}{c|p{1.5cm}p{1.5cm}|p{1.5cm}p{1.5cm}}
    \cline{1-5}
    \rule{0pt}{1\normalbaselineskip}
    \multirow{2}{*}{Methods}  
    &\multicolumn{2}{|c}{Photo domain} 
    &\multicolumn{2}{|c}{Unseen domain} \\ [0.5ex]
    \cline{2-5}
    
    \rule{0pt}{1\normalbaselineskip}
    &\multicolumn{1}{|c}{Content $\uparrow$}  
    &\multicolumn{1}{c}{Style $\uparrow$} 
    &\multicolumn{1}{|c}{Content $\uparrow$}  
    &\multicolumn{1}{c}{Style $\uparrow$}\\ [0.5ex]
    \hline \hline
    
    \rule{0pt}{1\normalbaselineskip}
    DiffusionCLIP   &\hfil 3.71   &\hfil 2.95   &\hfil 3.29   &\hfil 3.05 \\
    Ours    &\hfil \bf 4.76    &\hfil \bf 4.71  &\hfil \bf 4.62   &\hfil \bf 4.71\\ [0.5ex]
    \cline{1-5}
\end{tabular}
}
\vspace{0.1cm}
\caption{User study results on comparison with DiffusionCLIP.}

\label{table_diffusion}
\end{table}

In contrast, our proposed method maintains structural information and retains hats and hands in the outputs, without crushing any details of the face, such as hairs or eyes. Additionally, our method generates outputs with feasible texture, unlike StyleCLIP which produces outputs that still look like photos, or StyleGAN-NADA which struggles to translate images into pop art style. Similarly, VQGAN-CLIP and CLIPstyler fail to generate Pixar and Uiyo-e style images, while our method provides high-fidelity samples transferred into the styles of the target prompt.

Our proposed method's superiority is further supported by quantitative evaluation. As shown in Table~\ref{table_gan}, our method achieves the highest scores in both user study and CLIP score. While StyleCLIP obtains the smallest face identity loss, this suggests that it preserves semantic information to such an extent that it fails to transform the style adequately. On the other hand, VQGAN-NADA and CLIPstyler over-modulate the images, resulting in significant content alteration. In contrast, our method achieves a balance between content preservation and style transfer. The CLIP score is calculated globally, as described in equation \eqref{eq_clip_glob_loss}, and in a patch-based manner. Face identity loss is measured using ArcFace~\cite{deng2019arcface}. The same images used in the user study are used for the quantitative experiments.

\begin{figure}[!t]
	\centerline{\includegraphics[width=0.9\columnwidth]{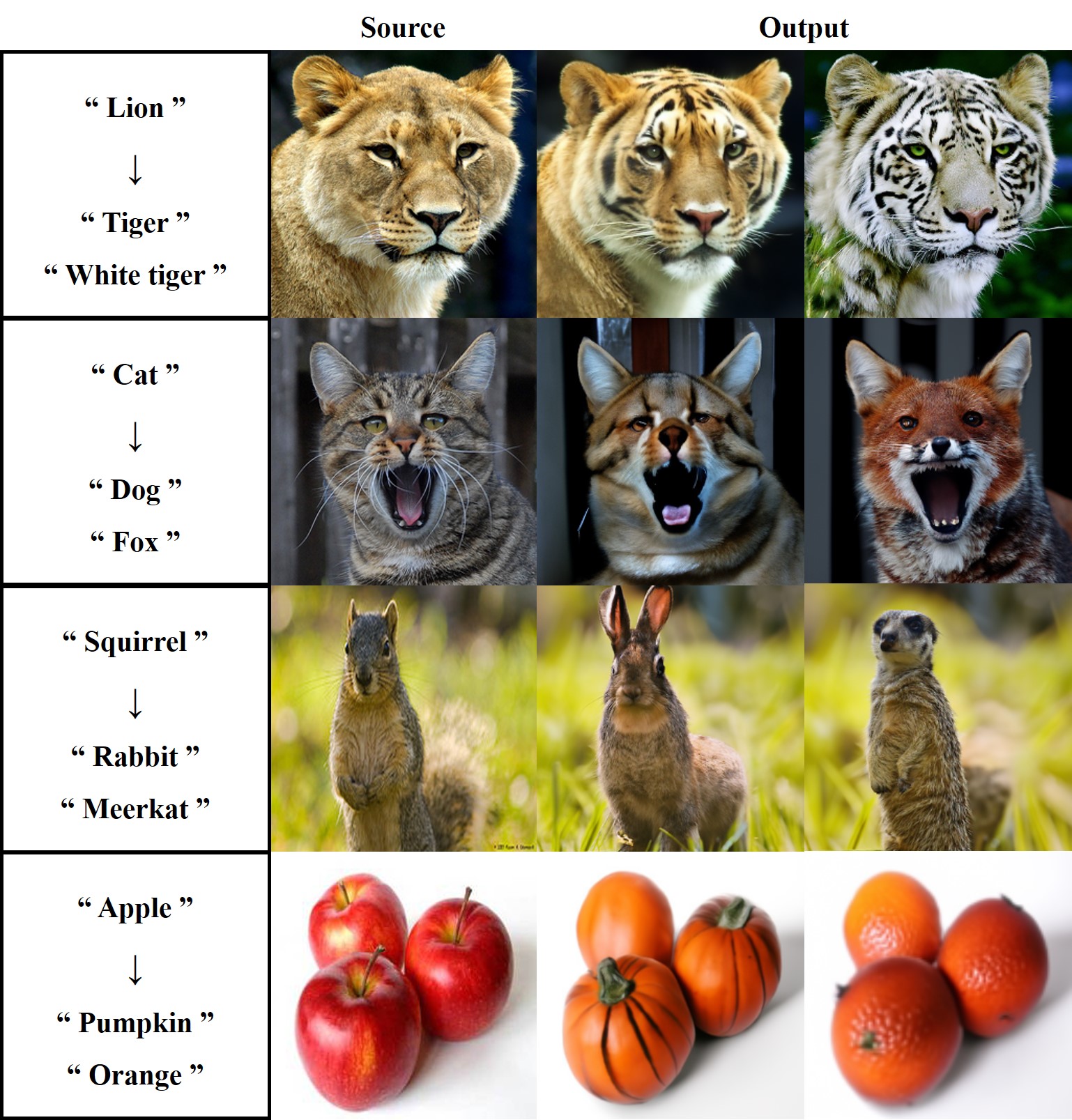}}
	\caption{Image translation results.}
	\label{fig_result_image_translation}
\end{figure}

\paragraph{Comparison with diffusion models}
We compared our proposed method with three diffusion-based models, DDIB~\cite{su2022dual}, DiffusionCLIP~\cite{kim2022diffusionclip}, and DiffuseIT~\cite{kwon2022diffusion}. To evaluate the translation performance of DDIB between painting and photo domains, we trained a new diffusion model on the Wikiart dataset. For the photo domain, we used the pretrained diffusion models described above. The qualitative and quantitative results of the comparison are presented in Figure \ref{fig_result_diffusion} and Table~\ref{table_diffusion}.

The third row of Figure \ref{fig_result_diffusion} shows that DDIB suffers from identity loss, where the facial identity of the portrait is destroyed in the translation from portrait to photo domain. Moreover, the shape of the church is not well delineated in the output of DDIB. Additionally, diffusion models have to be trained for each new domain, which is a critical drawback of DDIB. Therefore, image translation from portrait to neon light style is not available with DDIB.

On the other hand, DiffusionCLIP shows relatively satisfying quality in translating photos into another style. However, when the input images are not well converted into the photo domain, the results from unseen domain images are unsatisfactory, as shown in the first and third columns of Figure \ref{fig_result_diffusion}. This is supported by user study results on DiffusionCLIP, as presented in Table \ref{table_diffusion}, where the content score in unseen domains is 0.42 lower than the score in the photo domain. 
Furthermore, DiffuseIT shows the trade-off between style transfer and content preservation as shown in the fifth row of Figure \ref{fig_result_diffusion}. While changing the style of the source image, the facial identity is also modified. As demonstrated in the last column, the neon light is hardly seen when the shape of the church is well-preserved.

\begin{figure}[!t]
	\centerline{\includegraphics[width=\columnwidth]{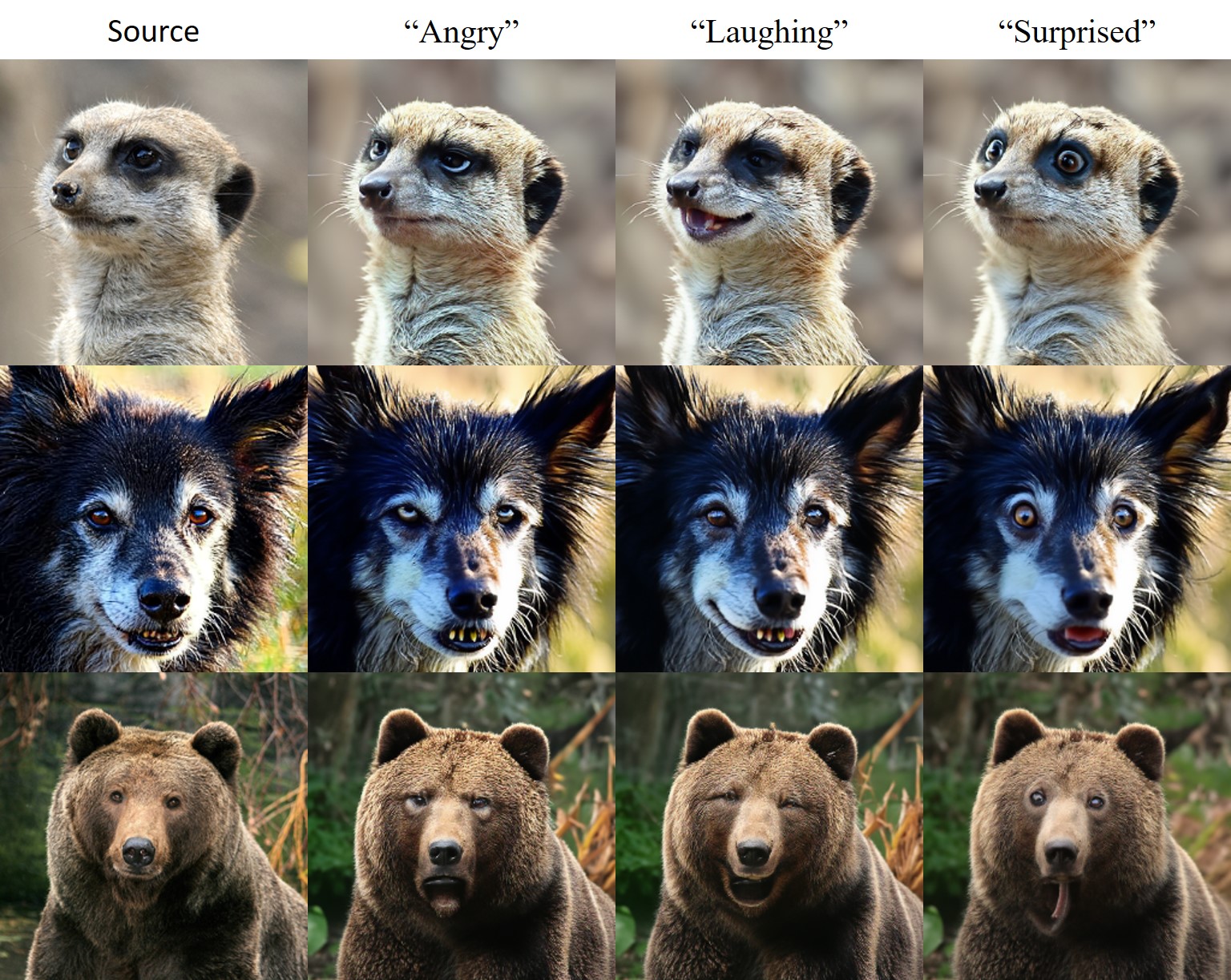}}
	\caption{Image manipulation results with emotional prompts. }
	\label{fig_result_image_editing}
\end{figure}

In contrast, our proposed method can stylize images not only from photo domains but also from unseen domains, such as portraits or paintings. The portrait is transformed into a photo while maintaining its facial identity, and the painting of a church is translated into neon light style while retaining small objects like a cross. These results are confirmed with user study results presented in Table \ref{table_diffusion}, where the scores between the photo domain and unseen domains are highly similar. This means that our method can modulate images even from unseen domains.
Regarding computational time, as shown in Table \ref{table_computation}, our method is significantly faster than DiffusionCLIP.
Additional examples are provided in the Supplementary material.

\begin{figure}[!t]
	\centerline{\includegraphics[width=1\columnwidth]{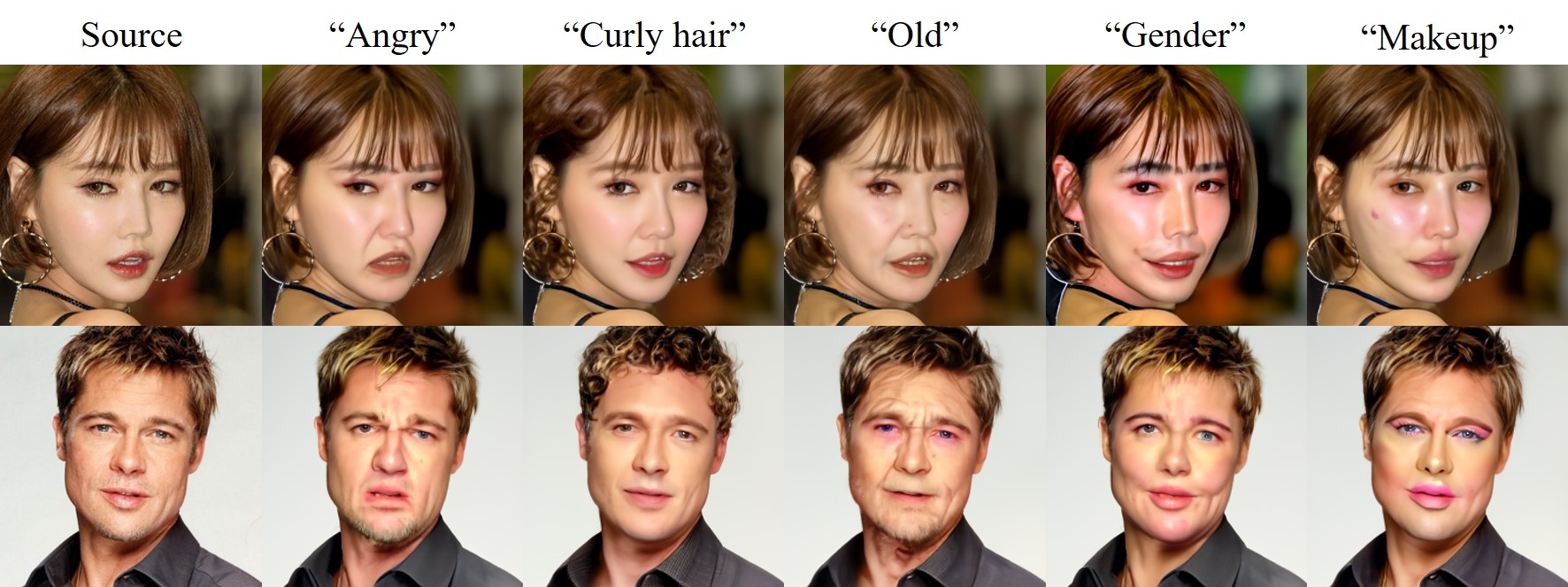}}
	\caption{Image manipulation results with human faces. }
	\label{fig_result_image_editing2}
\end{figure}

\paragraph{Image manipulation}

Our proposed method not only excels in image style transfer but also has potential for other tasks such as simple image translation and manipulation. The qualitative results for image translation are shown in Figure \ref{fig_result_image_translation}, where our method can translate different animal species while preserving the details and maintaining the overall coherence of the image. In addition, our method can also change from apples to pumpkins or oranges, as shown in the second row of Figure \ref{fig_result_image_translation}.

Moreover, our method can also be used for image manipulation tasks, as demonstrated in Figure \ref{fig_result_image_editing} and Figure \ref{fig_result_image_editing2}. Figure \ref{fig_result_image_editing} shows an example of changing the expression of animals, while Figure \ref{fig_result_image_editing2} shows an example of appearance manipulation such as age, gender and make-up. These results demonstrate the potential of our method for various image manipulation tasks.

\begin{table}[!t]
\centering
\resizebox{\columnwidth}{!}{
\begin{tabular}{c|p{2.5cm}p{2.5cm}p{2cm}p{3cm}}
    \hline
    \rule{0pt}{1\normalbaselineskip}
    Methods  &\hfil Data preparation  &\hfil \# Train Param.  &\hfil Training time   &\hfil Inference time (sec) \\ [0.5ex]   
    \hline
    \rule{0pt}{1\normalbaselineskip}
    ILVR    &\hfil -   &\hfil -    &\hfil -    &\hfil 100   \\
    DDIM    &\hfil -   &\hfil -    &\hfil -    &\hfil 11    \\
    DDIB    &\hfil -   &\hfil 1104 M   &\hfil $>$ 200 hrs  &\hfil 12    \\
    DiffusionCLIP   &\hfil 5.85 min &\hfil 113 M    &\hfil 293 sec    &\hfil 96    \\
    DiffuseIT   &\hfil- &\hfil -   &\hfil -    &\hfil 40    \\
    Ours    &\hfil -   &\hfil -    &\hfil -    &\hfil 38    \\ [0.5ex]
    \hline
\end{tabular}
}
\vspace{0.1cm}
\caption{Comparison on computational complexity of various diffusion models. The symbol ``-'' indicates that data preparation or training is not required. }
\label{table_computation}
\end{table}

\begin{figure*}[!hbt]
	\centerline{\includegraphics[width=0.9\textwidth]{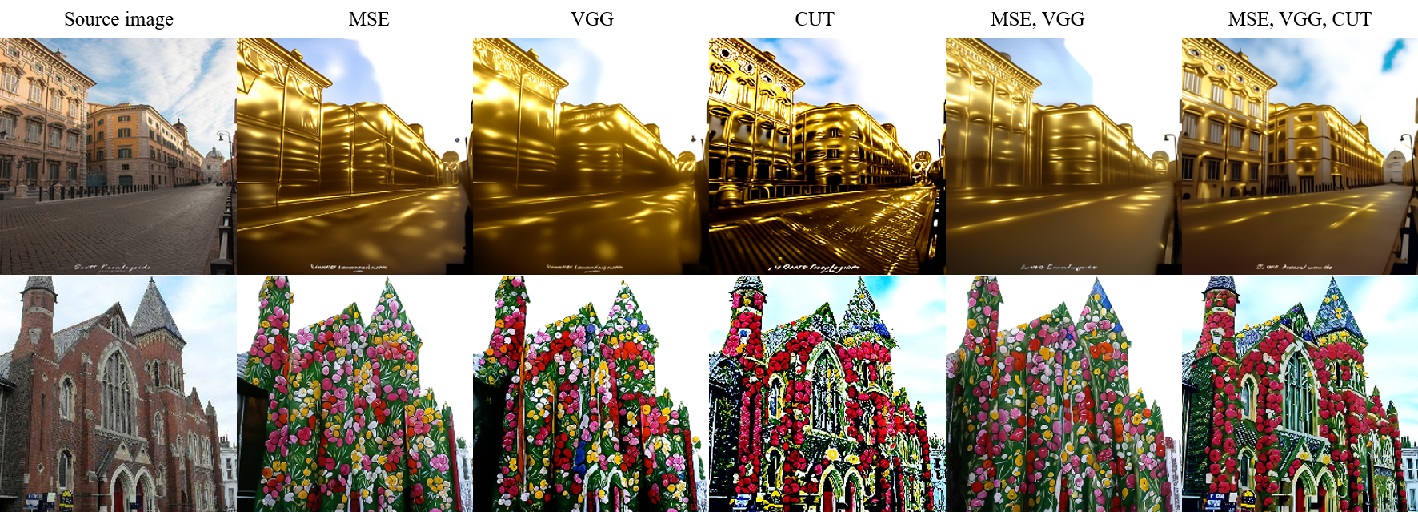}}
	\caption{
	Ablation study focusing on three losses for content guidance - $\ell_{MSE}$, $\ell_{VGG}$, and $\ell_{ZeCon}$. The results show that, in each row of translated images, which are transformed into the styles of ``golden" and ``oil painting of flowers", the proposed patch-wise content preservation loss is effective in preserving content information.
	% In each row, source images are translated into the style of "golden" and "oil painting of flowers", respectively. Patch-wise contrastive loss helps to preserve content information.
	}
    \vspace{-0.5cm}
	\label{fig_Loss_Content}
\end{figure*}

\begin{figure}[!hbt]
	\centerline{\includegraphics[width=0.9\columnwidth]{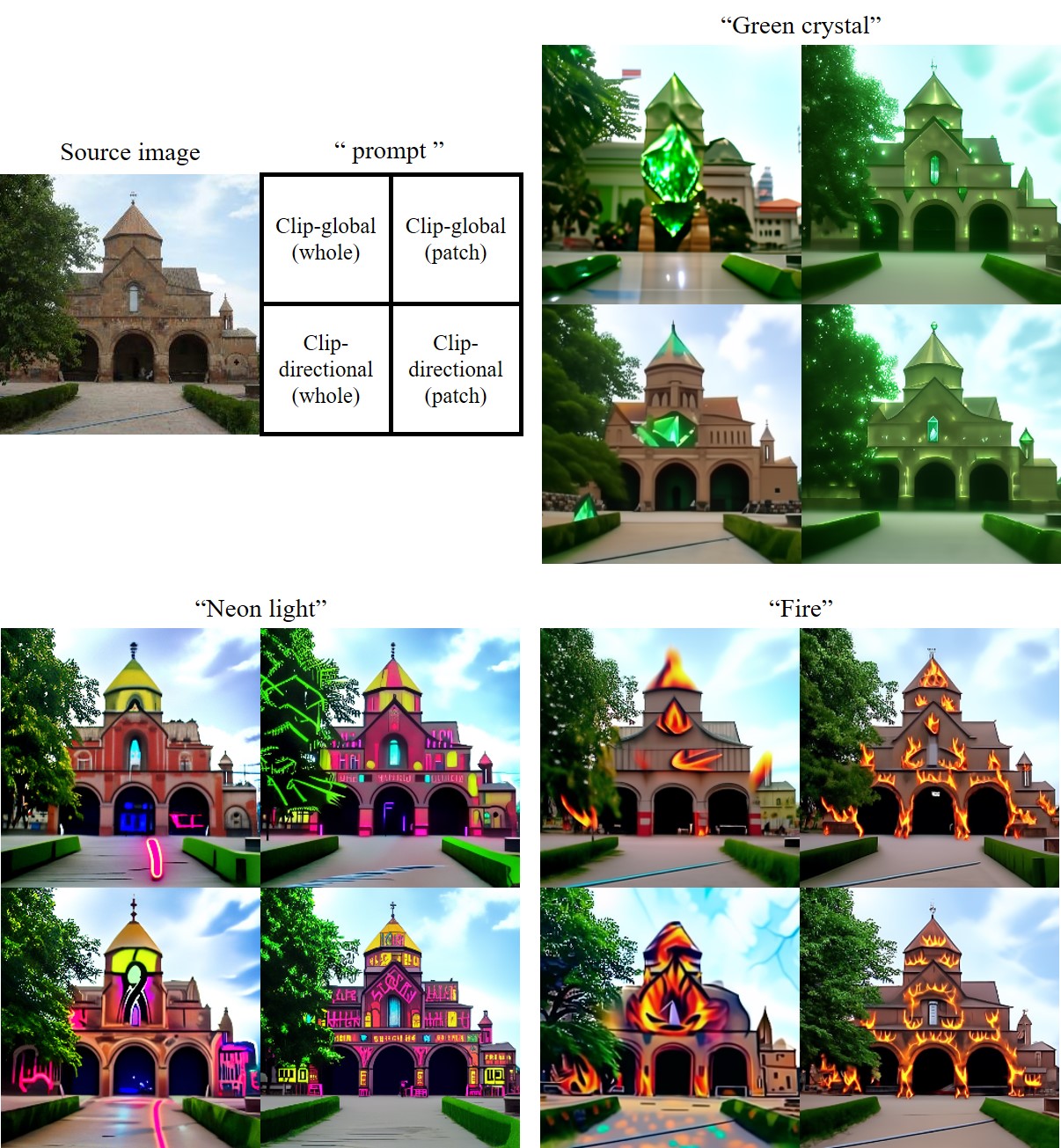}}
	\caption{Ablation study on four losses for style guidance - CLIP global loss, patch-based CLIP global loss, CLIP directional loss, and patch-based CLIP directional loss. }
	\label{fig_Loss_Style}
\end{figure}

\subsection{Ablation studies}

\paragraph{Roles of content losses}
To investigate the effectiveness of content guidance losses, we performed ablation studies. Our proposed content loss for guidance in \eqref{eq_guidance_content} consists of three different losses, namely $\ell_{ZeCon}$, $\ell_{VGG}$, and $\ell_{MSE}$. To examine the contribution of $\ell_{ZeCon}$, we eliminated it from the total content loss and compared the results with the complete content loss. In Figure \ref{fig_Loss_Content}, we observed that excluding $\ell_{ZeCon}$ resulted in a loss of structural details such as windows in the building outlines, even though the overall shape was preserved. This suggests that $\ell_{VGG}$ and $\ell_{MSE}$ alone are insufficient to preserve the fine-grained details of the content.

On the other hand, employing all three losses yielded the best results in terms of content preservation. The user study results, presented in Table \ref{table_ablation_cut}, support the superiority of $\ell_{ZeCon}$ compared to $\ell_{MSE}$, $\ell_{VGG}$, and even the combination of both. This implies that $\ell_{ZeCon}$ effectively preserves the structural details while avoiding over-fitting. In summary, the ablation studies demonstrate the crucial role of $\ell_{ZeCon}$ in our proposed method for preserving the structural properties of the input images.

\paragraph{Roles of style losses}

We conducted ablation studies to investigate the role of each loss function in our style loss. The loss function consists of two parts, $\ell_{global}$ and $\ell_{dir}$, as shown in  \eqref{eq_guidance_clip}. To examine the contribution of each loss, we applied the loss functions individually and evaluated the results on three different styles - green crystal, neon light, and fire. The qualitative results are shown in Figure \ref{fig_Loss_Style}. We found that applying the directional CLIP loss in addition to the global CLIP loss  led to more stylized images, as compared to applying the global CLIP loss alone. This implies that directional CLIP loss is more effective in modulating the style of images.

We also verified the role of patch-based guidance in our proposed method. We observed that using whole-image guidance tends to stylize the image in local parts, while patch-based guidance transforms the image into the given style by covering a large area. This is demonstrated in Figure \ref{fig_Loss_Style}, where the patch-based guidance is applied to stylize the entire background while preserving the foreground object.

Overall, the results of our ablation studies suggest that both $\ell_{global}$ and $\ell_{dir}$ are important for achieving high-quality style transfer results, and that patch-based guidance is effective in transforming images into a given style.

\begin{table}[!t]
\centering
\resizebox{0.6\columnwidth}{!}{
\begin{tabular}{c|cc}
    \hline
    \rule{0pt}{1\normalbaselineskip}
    &\multicolumn{1}{c}{Content $\uparrow$}  
    &\multicolumn{1}{c}{Style $\uparrow$}   \\  [0.5ex] 
    \hline\hline
    
    \rule{0pt}{1\normalbaselineskip}
    MSE   &\hfil 3.29   &\hfil 3.81 \\
    VGG   &\hfil 3.00   &\hfil 3.86 \\ 
    ZeCon   &\hfil \underline{4.29}   &\hfil \underline{4.57} \\
    MSE, VGG   &\hfil 3.29   &\hfil 3.95 \\
    MSE, VGG, ZeCon   &\hfil \bf 4.81   &\hfil \bf 4.81 \\ [0.5ex] 
    \hline  
\end{tabular}
}
\vspace{0.2cm}
\caption{User study results on ablation studies regarding content losses. Bold text and underline refer to the best and the second best scores, respectively.}

\label{table_ablation_cut}
\end{table}

\section{Conclusion}

In this paper, we proposed a novel method for diffusion-based image style transfer without content changes using Zero-Shot Contrastive (ZeCon) loss. One of the major advantages of our method is that it is training-free and does not require additional training or data, which significantly reduces the computational time. Our experiments showed that contrastive loss with diffusion model leads to high capability in maintaining content while achieving effective stylization. Additionally, our method demonstrated potential for image translation and manipulation tasks. The limitations of our method are discussed in the Supplementary material.

%\clearpage
%-------------------------------------------------------------------------

{\small
\bibliographystyle{ieee_fullname}
\bibliography{egbib}
}

\clearpage

\appendix

{\noindent \Large \bf {Supplementary Material}}

\begin{figure*}[!hbt]
	\centerline{\includegraphics[width=1.7\columnwidth]{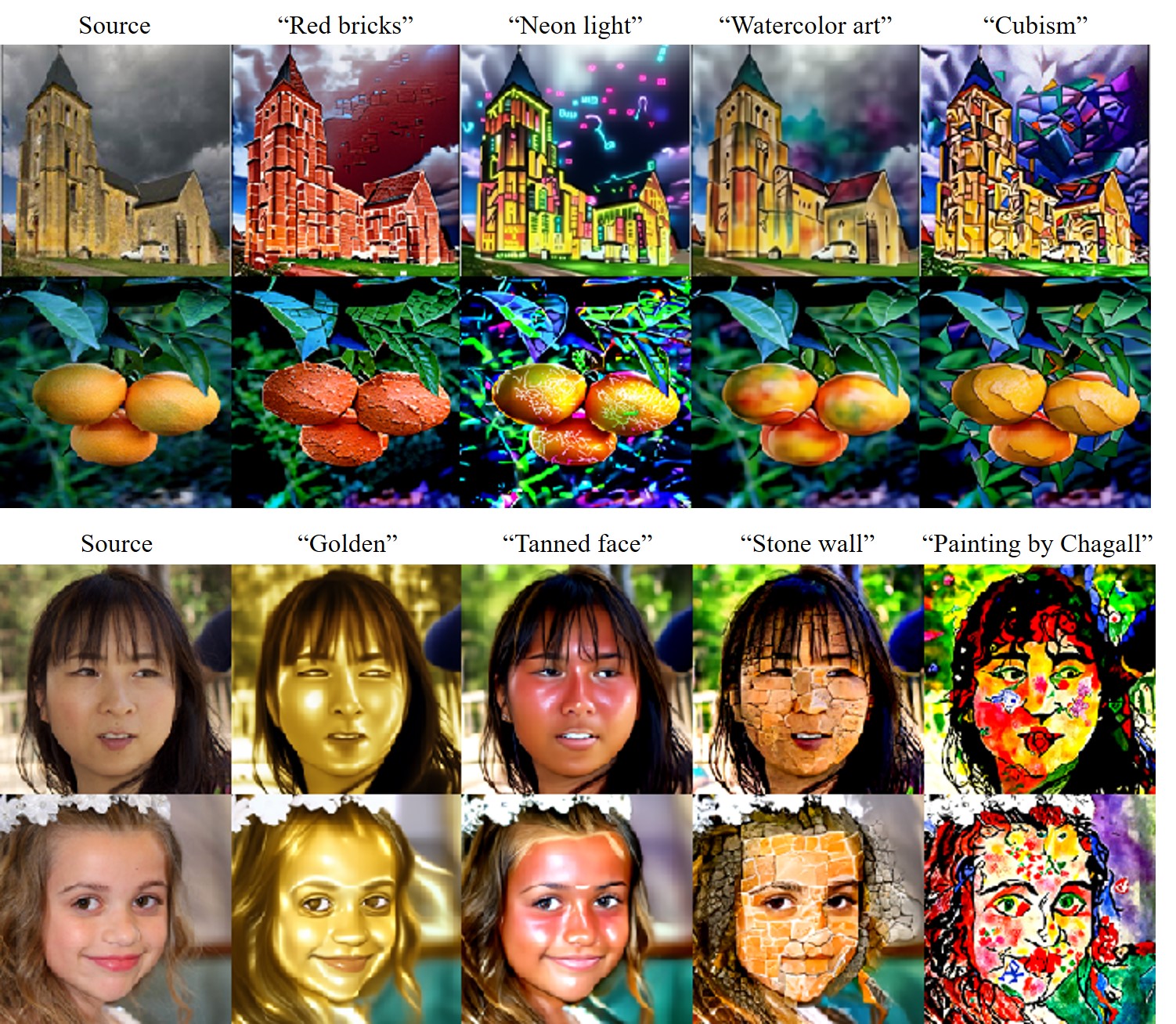}}
	\caption{Additional results on various style prompts. }
	\label{fig_supple_more2}
\end{figure*}

\section{Details of Implementation}

\subsection{Data manipulation and hyperparameters}
\label{appendix_data}
To apply directional CLIP loss in a patch-based manner~\cite{kwon2022clipstyler}, we start by randomly cropping the source image. The patch size can vary, but for texture styles like golden or green crystal, we mainly use $(0.01, 0.05)$, while for artistic styles like painting by Gogh or pop art, we use $(0.01, 0.3)$. The cropped images are then augmented with perspective function and random affine transformation.

To guide both style and content, we use different weights for each loss according to the style. Although the values may differ for each style, the weights for $L_{global}$ and $L_{dir}$ usually range from 5000 to 30000. Additionally, Table \ref{table_lambda} provides examples of the weights used for $L_{ZeCon}$, $L_{VGG}$, and $L_{MSE}$. However, one can adjust these values to improve the quality of the final image.

\subsection{Patch-wise cross entropy loss for ZeCon guidance}
\label{appendix_cut}
We provide a more detailed explanation of the cross-entropy loss in equation \eqref{eq_CUT_loss}. The loss takes a query patch $\boldsymbol{v}$, along with its positive counterpart $\boldsymbol{v^{+}}$ and $N$ negative counterparts $\boldsymbol{v}^{-}_{i}$, where $i \in [1, \dots, N]$, as inputs. The query patch is taken from the generated image, while the positive patch is the corresponding patch from the source image. The negative patches are non-corresponding patches from the source image.
The purpose of the cross-entropy loss is to encourage a patch to share the embedding space with its corresponding patch from the input, and not with the other patches. Mathematically, the cross-entropy loss can be expressed as:
\begin{equation}
\ell(\boldsymbol{v},\boldsymbol{v^{+}},\boldsymbol{v^{-}}) = -log \left[ \frac{e^{\boldsymbol{v} \cdot \boldsymbol{v^{+}}/\tau}}{e^{\boldsymbol{v} \cdot \boldsymbol{v^{+}}/\tau}+\Sigma_{i=1}^{N}e^{\boldsymbol{v} \cdot \boldsymbol{v^{-}}_i/\tau}} \right]
\end{equation}
where $\tau$ is a temperature.
% implementation. 

\subsection{Training DDIB}
To compare with DDIB, we trained a diffusion model using 13,000 images from the Wikiart dataset, each with dimensions of $256 \times 256$. The model architecture is based on guided diffusion~\cite{dhariwal2021diffusion}, with $128$ base channels and attention at $16 \times 16$ and $8 \times 8$ resolutions. We did not use residual blocks for upsampling and downsampling, and we fixed the variance as a constant~\cite{ho2020denoising}. The model was trained for 50,000 iterations using a batch size of $8$ on an NVIDIA RTX 3090.

\subsection{Details of image manipulation}
While patch-based CLIP losses are effective for modulating textures, they are not ideal for image translation tasks that involve translating entire classes of objects. For these tasks, guidance must be applied to the entire image rather than just small patches. Therefore, in Figures \ref{fig_result_image_translation}, \ref{fig_result_image_editing}, and \ref{fig_result_image_editing2}, CLIP losses were calculated for the entire image in both image translation and manipulation tasks. However, for the image style transfer task, patch-based CLIP losses were utilized, as they are well-suited to modulating texture styles.

\begin{figure*}[!t]
	\centerline{\includegraphics[width=0.9\textwidth]{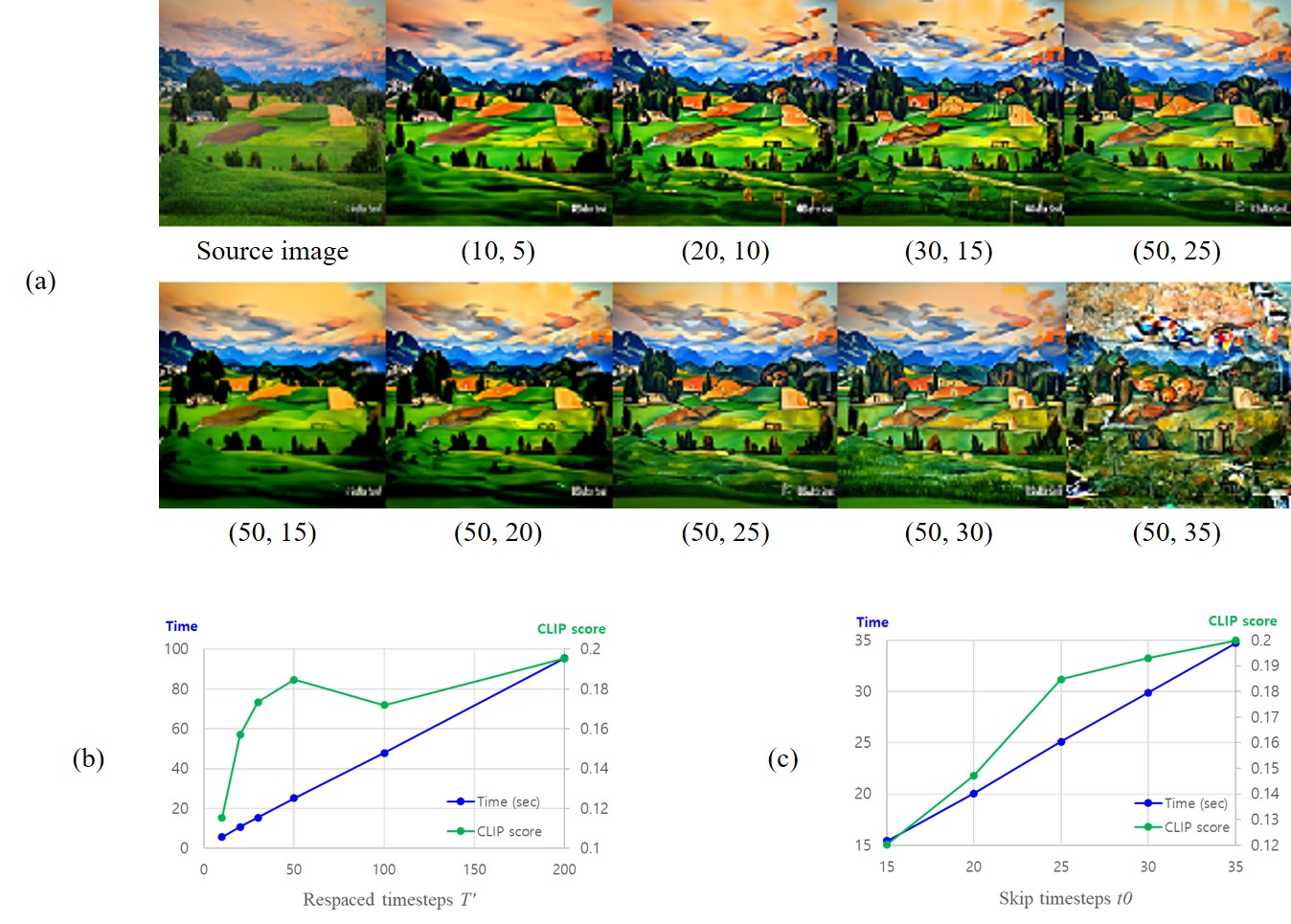}}
	\caption{The effect of the respaced time $T'$ and skipped time $t_0$. (a) demonstrates the images sampled with ($T'$,$t_0$). The first row shows the difference between various $T'$ when $t_0$ is its half and the second row shows the difference between various $t_0$ when $T' = 50$. (b) and (c) illustrates the relationship between sampling time and CLIP score as graphs for the first and second rows of (a), respectively.}
	\label{fig_supple_timesteps}
\end{figure*}

\section{Additional experimental results}

\subsection{Effect of the number of timesteps}
\label{section_ts}
Since the diffusion process usually takes lots of time, two techniques are widely used - respacing and skipping time steps~\cite{chung2022come, kim2022diffusionclip}. The last time step $T$ is respaced into $T'$. Then we forward the diffusion model to time $t_0 < T'$ and reverse the diffusion process from $x_{t_0}$. $T'$ and $t_0$ have various effects on both image quality and time consumption. As shown in the Figure \ref{fig_supple_timesteps} (a) and (b), image quality with respect to style transformation enhances as respacing time step $T'$ increases. However, its growth rate decreases and its difference is imperceptible even though sampling time still increases. 
In the mean time, CLIP score increases as $t_0$ increases as illustrated in Figure \ref{fig_supple_timesteps} (c). On the other hand, the content information is not fully preserved in the time steps $t_0 = 15$ or $20$ as shown in the Figure \ref{fig_supple_timesteps} (a). Thus, we set ($T'$, $t_0$) as ($50$, $25$) for our baseline.

\subsection{Diffusion models' trade-off between style and content}

With respect to style transfer, one of the challenges posed by unconditional diffusion models is to maintain content of the given image. When transforming styles of the given image, its content changes simultaneously. GAN-based methods explicitly impose content losses, such as a reconstruction loss. This results in good performance in content preservation. In contrast, diffusion models have no constraint during training phase. They generate high quality images in correspondence with the training data domain. The semantic constraints are not considered which finally results in the degradation in the quality of the generated images. 

Here, we compare four diffusion models - ILVR, DDIM, DiffusionCLIP, and our proposed method - with respect to style and content in the Figure \ref{fig_schemes}. ILVR utilizes down-sampled reference image as condition in each reverse denoising steps. The condition helps the generated image share its content information with the reference image. However, it cannot have same identity because reverse DDPM steps without condition should be given sufficiently in order to generate images in photo style. This accordingly results in a loss of content. DDIM can reconstruct the source image when the variance of noise $\sigma_t$ is set as $0$. However, the style is also preserved with zero variance. When we control $\sigma_t$ as larger than zero, we can get photo style images but their content is 
altered with stochastic noise. DiffusionCLIP tried to solve the trade-off between content and style by fine-tuning the score function $\epsilon_{\theta}$. However, it requires much more time due to the model training for each style and data preparation. In addition, the content cannot be maintained when the source images are from unseen domains. In contrast, our proposed loss $\mathcal{L}_{ZeCon}$ does not require additional networks or fine-tuning. This leads to shorter time than DiffusionCLIP. With the help of ZeCon guidance, we could retain the content of source image from any domain and translate it into different styles.

\subsection{More Comparison with GAN and CNN-based methods}
\begin{figure*}[t]
	\centerline{\includegraphics[width=0.9\textwidth]{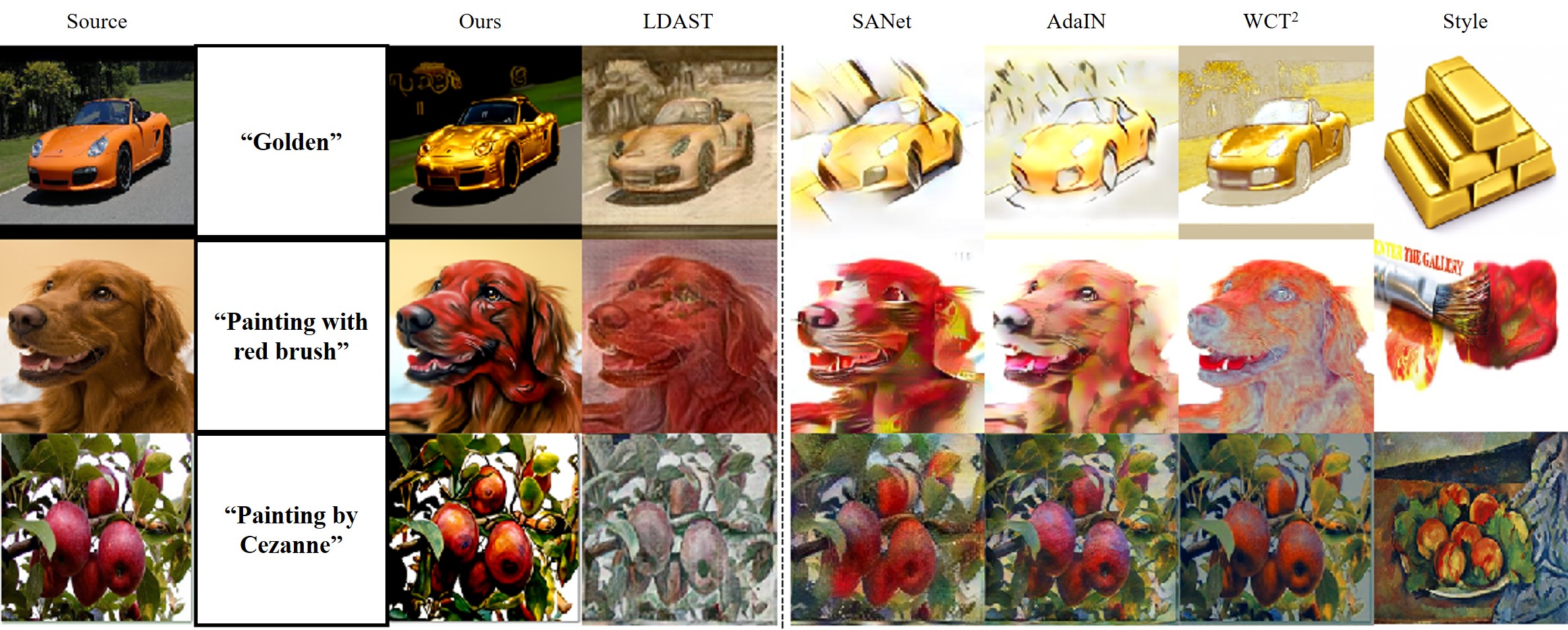}}
	\caption{Comparative study results. }
	\label{fig_supple_comparison_more}
\end{figure*}

In addition to the comparative studies on the GAN-based methods in the Section \ref{section_GAN}, we conducted more comparisons with various GAN-based and CNN-based methods including both text and image guidance. For text guidance method, we compared our method with one more method, LDAST~\cite{fu2022language}. For image guidance, we included three methods, SANet~\cite{fan2017sanet}, AdaIN~\cite{huang2017arbitrary}, and WCT2~\cite{yoo2019photorealistic}. As shown in the Figure  \ref{fig_supple_comparison_more}, we could notice that LDAST and WCT2 could preserve the content information better than SANet and AdaIN. However, all the four methods for comparison show inferior performance than our method in perspective of style transformation.

\begin{figure*}[!t]
	\centerline{\includegraphics[width=1.7\columnwidth]{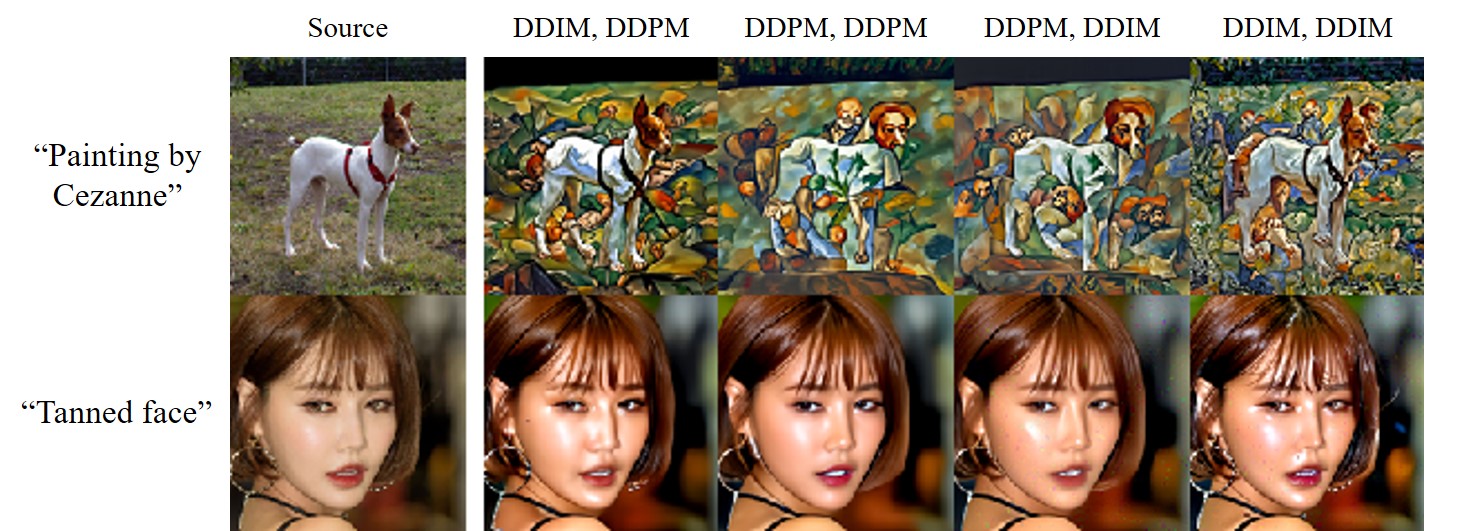}}
	\caption{Ablation study results on diffusion processes. From the second column to the right, the combinations of methods (forward, reverse) are (DDIM, DDPM), (DDPM, DDPM), (DDPM, DDIM), and (DDIM, DDIM).}
	\label{fig_supple_forward_reverse}
\end{figure*}

\begin{figure}[!h]
	\centerline{\includegraphics[width=\columnwidth]{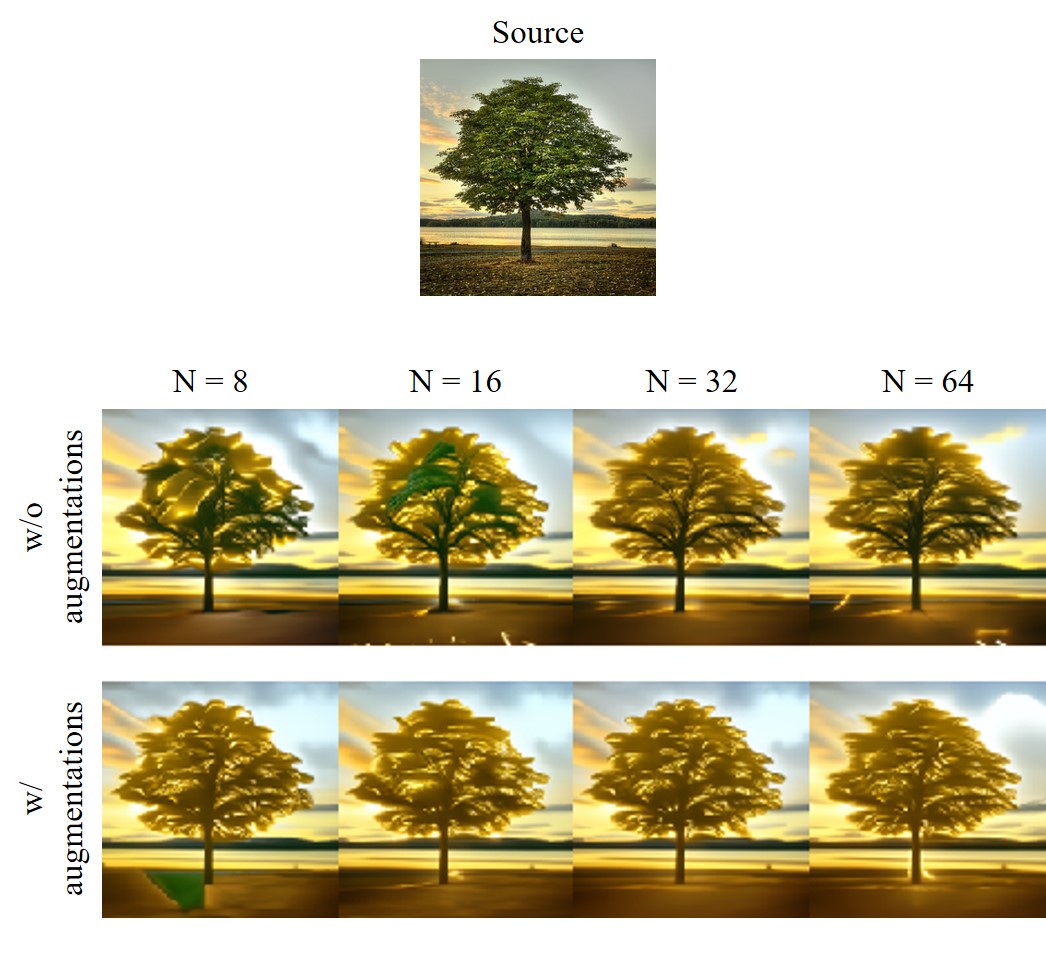}}
	\caption{Ablation study results on augmentation and the number of patches $N$. The target prompt is ``Golden."}
	\label{fig_supple_augmentation}
\end{figure}

\begin{figure}[!h]
	\centerline{\includegraphics[width=\columnwidth]{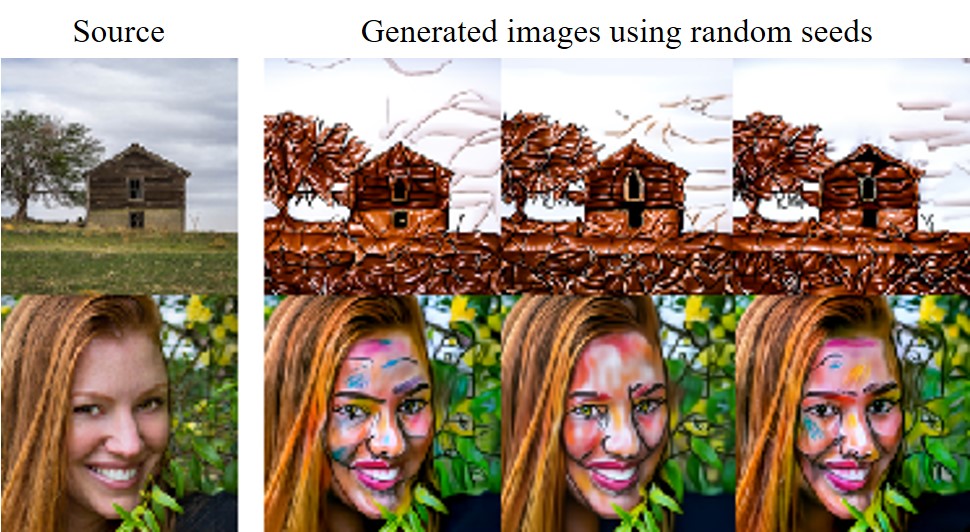}}
	\caption{Generated images using random seeds. The style prompts of the first and the second rows are ``Leather" and ``A sketch of crayon", respectively}
	\label{fig_supple_seed}
\end{figure}

\begin{figure}[!h]
	\centerline{\includegraphics[width=\columnwidth]{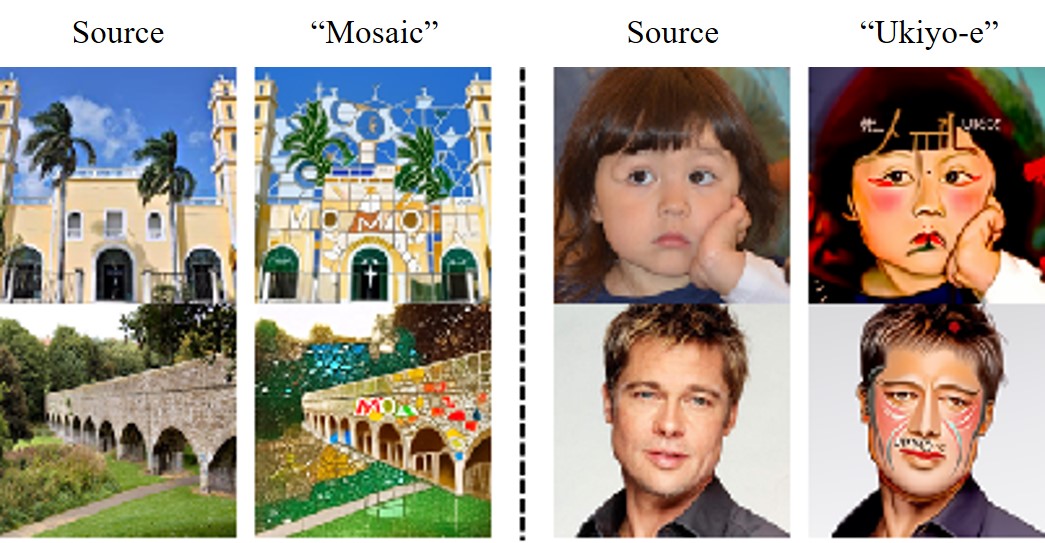}}
	\caption{Failure cases. Texts from target prompts sometimes appear on the generated images.}
	\label{fig_supple_limitation}
\end{figure}

\begin{table*}[!ht]
\centering
\resizebox{0.9\textwidth}{!}{
\begin{tabular}{p{2cm}|p{5cm}|p{3cm}p{3cm}|p{3cm}p{3cm}p{3cm}|p{1.5cm}p{1.5cm}}
\hline
\rule{0pt}{1\normalbaselineskip}
\hfil Model    &\hfil Style prompt  &\hfil CLIP-global  &\hfil CLIP-directional  &\hfil ZeCon   &\hfil MSE   &\hfil VGG   &\hfil Patch size   &\hfil $t_0$ \\ [0.5ex]   
\hline
\rule{0pt}{1\normalbaselineskip}
\multirow{10}{*}{\hfil ImageNET}
& \hfil Golden                      &\hfil 5000    &\hfil 5000    &\hfil 100     &\hfil 5000  &\hfil 10     &\hfil 0.05  &\hfil 15  \\
& \hfil Watercolor art              &\hfil 5000    &\hfil 10000    &\hfil 300      &\hfil 0      &\hfil 100    &\hfil 0.3  &\hfil 25  \\
& \hfil Stained glasses            &\hfil 15000    &\hfil 15000    &\hfil 200      &\hfil 1000  &\hfil 10    &\hfil 0.05   &\hfil 25  \\
& \hfil Oil painting of flowers     &\hfil 20000    &\hfil 20000    &\hfil 1500     &\hfil 10000  &\hfil 10     &\hfil 0.05  &\hfil 25  \\
& \hfil Red bricks                  &\hfil 20000    &\hfil 40000    &\hfil 1000     &\hfil 1000   &\hfil 10     &\hfil 0.05  &\hfil 25  \\
& \hfil Wooden                      &\hfil 20000    &\hfil 50000    &\hfil 1000      &\hfil 1000  &\hfil 10    &\hfil 0.05  &\hfil 25  \\
& \hfil Leather                     &\hfil 20000    &\hfil 30000    &\hfil 2000     &\hfil 20000   &\hfil 200     &\hfil 0.3  &\hfil 25  \\
& \hfil Marbling                    &\hfil 20000    &\hfil 30000    &\hfil 2000     &\hfil 20000  &\hfil 200    &\hfil 0.3   &\hfil 25  \\
& \hfil Autumn                      &\hfil 20000    &\hfil 20000    &\hfil 700      &\hfil 10000  &\hfil 100    &\hfil 0.05  &\hfil 25  \\
& \hfil Snowy                       &\hfil 20000    &\hfil 20000    &\hfil 700      &\hfil 0  &\hfil 100    &\hfil 0.05  &\hfil 25  \\
\hline
\multirow{10}{*}{\hfil FFHQ}
& \hfil Pop art                         &\hfil 10000    &\hfil 20000    &\hfil 50      &\hfil 1000    &\hfil 50    &\hfil 0.3  &\hfil 25  \\
& \hfil Stone wall                      &\hfil 2000    &\hfil 50000    &\hfil 500     &\hfil 5000  &\hfil 10     &\hfil 0.1  &\hfil 25  \\
& \hfil Tanned face                      &\hfil 15000    &\hfil 15000    &\hfil 1000     &\hfil 10000   &\hfil 100     &\hfil 0.3  &\hfil 25  \\
& \hfil Clay                            &\hfil 40000    &\hfil 40000    &\hfil 1000     &\hfil 10000  &\hfil 0    &\hfil 0.05   &\hfil 25  \\
& \hfil Portrait by Gogh                &\hfil 10000    &\hfil 7000    &\hfil 10     &\hfil 3000   &\hfil 50     &\hfil 0.3  &\hfil 25  \\
& \hfil A sketch with crayon            &\hfil 10000    &\hfil 20000    &\hfil 500     &\hfil 10000   &\hfil 100     &\hfil 0.3  &\hfil 25  \\
& \hfil 3d render in the style of Pixar &\hfil 5000    &\hfil 5000    &\hfil 500     &\hfil 10000  &\hfil 100     &\hfil 0.3  &\hfil 25  \\
& \hfil Golden                         &\hfil 7000    &\hfil 7000    &\hfil 200      &\hfil 0  &\hfil 50    &\hfil 0.05   &\hfil 15  \\

& \hfil Ukiyo-e                 &\hfil 8000    &\hfil 20000    &\hfil 1000      &\hfil 5000  &\hfil 100    &\hfil 0.3  &\hfil 25  \\
& \hfil Marbling                &\hfil 20000    &\hfil 40000    &\hfil 1000      &\hfil 10000      &\hfil 10    &\hfil 0.3  &\hfil 25  \\
\hline
\end{tabular}
}
\vspace{0.2cm}
\caption{Examples of hyperparameters for various style prompts. Weights for CLIP-global loss, CLIP-directional loss, ZeCon loss, MSE loss, and VGG loss are given. For patch-based CLIP guidance, we control the patch size. The maximum size is given in the table with the minimum of 0.01. $t_0$ is the time step to which the source image is forwarded when $T'=50$.}
\label{table_lambda}
\end{table*}

\subsection{DDPM and DDIM for diffusion processes}
Although either DDPM or DDIM can be utilized for both forward and reverse processes, we conducted a comparative study  in order to show their differences in the generated images. As shown in Figure \ref{fig_supple_forward_reverse}, results from the forward DDIM show better performance in preserving content than DDPM. The earring in the second row appears unchanged in the output images generated by forward DDIM, whereas its shape is altered in the images produced by forward DDPM. For reverse process, DDPM tends to transform styles better compared to DDIM.
Accordingly,
we chose to use DDIM as forward and DDPM as reverse process as default.

\subsection{Unseen domains}
DiffusionCLIP tried to solve the trade-off between content and style by fine-tuning the diffusion model with identity loss. Because of the constraints imposed on the finetuned model, the transformed image shows high performance in identity preservation. However, the fine-tuned model $\hat{\epsilon}_{\theta}$ converts only the photo domain images. When it comes to unseen domains, such as portraits or paintings, they should be converted to photo images through $\epsilon_{\theta}$. Since it has not been fine-tuned with identity loss, the semantic information is lost during the reverse sampling process due to the stochastic property of the diffusion model. Thus, the final output from images of unseen domains is degraded in its quality.
In contrast, our proposed method can transform even the unseen domain images with only one step. Since our method can preserve the identity with content guidance, the final outputs do not suffer from quality degradation. Also, our method takes about 38 seconds whereas DiffusionCLIP requires about 400 seconds for model fine-tuning and sampling.
As described in the Figure \ref{fig_schemes}, DiffusionCLIP requires two steps from portrait to photo to Pixar domains. In this process, the face identity is destroyed. However, the proposed method could preserve the identity while transforming into the style of Pixar.

\subsection{Augmentation of patches}
We use patch-based CLIP losses for style guidance. From the denoised image $\hat{x}_{0,t}$, $N$ patches are randomly cropped and augmented with perspective function and random affine transformation. In order to check the importance of augmentation, we conducted ablation study on the augmentation and the number of patches. 
As shown in the Figure \ref{fig_supple_augmentation}, the images generated without augmentation could not transform to gold enough. 
Also, when $N < 32$, the tree is not sufficiently converted to align with the target prompt ``Golden". 
From these observations, we chose $N=32$ with augmentations and this choice results in further reduction in inference time to 24 seconds.

\subsection{Stochastic property}
The random nature of DDPM leads to various modifications generated from the same style prompt.
As shown in Figure \ref{fig_supple_seed}, we observed that the same image and text prompt pair could generate various images using different random seeds.

\subsection{User study}
For quantitative analysis, we conducted a user study. For comparison with GAN-based methods, 60 images with four styles have been used in total. The styles involved are ``golden", ``clay", ``3d render in the style of Pixar", and ``pop art." We utilized human face images because StyleCLIP and StyleGAN-NADA are based on face dataset. In addition, we totally generated 24 images with six styles for comparison with DiffusionCLIP. We chose three styles (``neon light", ``green crystal", and ``Ukiyo-e") for the photo domain and the other three styles (``3d render in the style of Pixar", ``pop art", and ``golden") for unseen domains. We used portraits and paintings from Wikiart dataset for unseen domain images. Besides, for ablation study on content losses, we used 15 images for three styles (``golden", ``oil painting of flowers", ``leather") in total. The number of questions were 14. 21 users participated in the user study. They were randomly recruited online.

\section{Limitations}
Although our proposed method has various strengths and shows great performance, there remain some limitations. As described in the Appendix \ref{appendix_data}, one should find weights for each loss though their relevant ranges are given in this paper. 
Also, it has been observed that in some cases, the text prompts describing the targeted style are displayed on the generated images (Figure \ref{fig_supple_limitation}).

%\clearpage

\begin{figure*}[!t]
	\centerline{\includegraphics[width=0.9\textwidth]{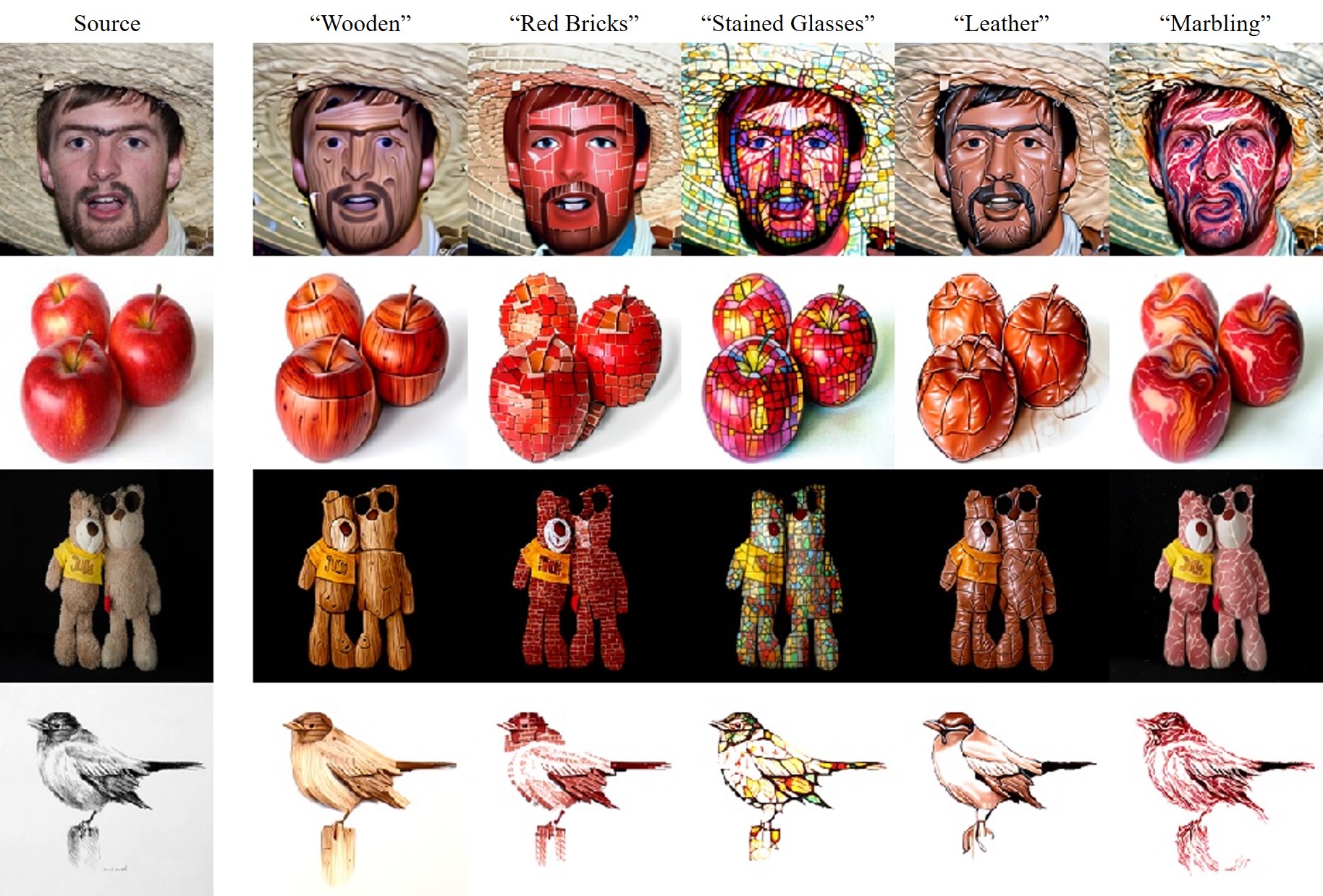}}
	\caption{Additional results on various style prompts. }
	\label{fig_supple_more}
\end{figure*}

\begin{figure*}[!t]
	\centerline{\includegraphics[width=0.9\textwidth]{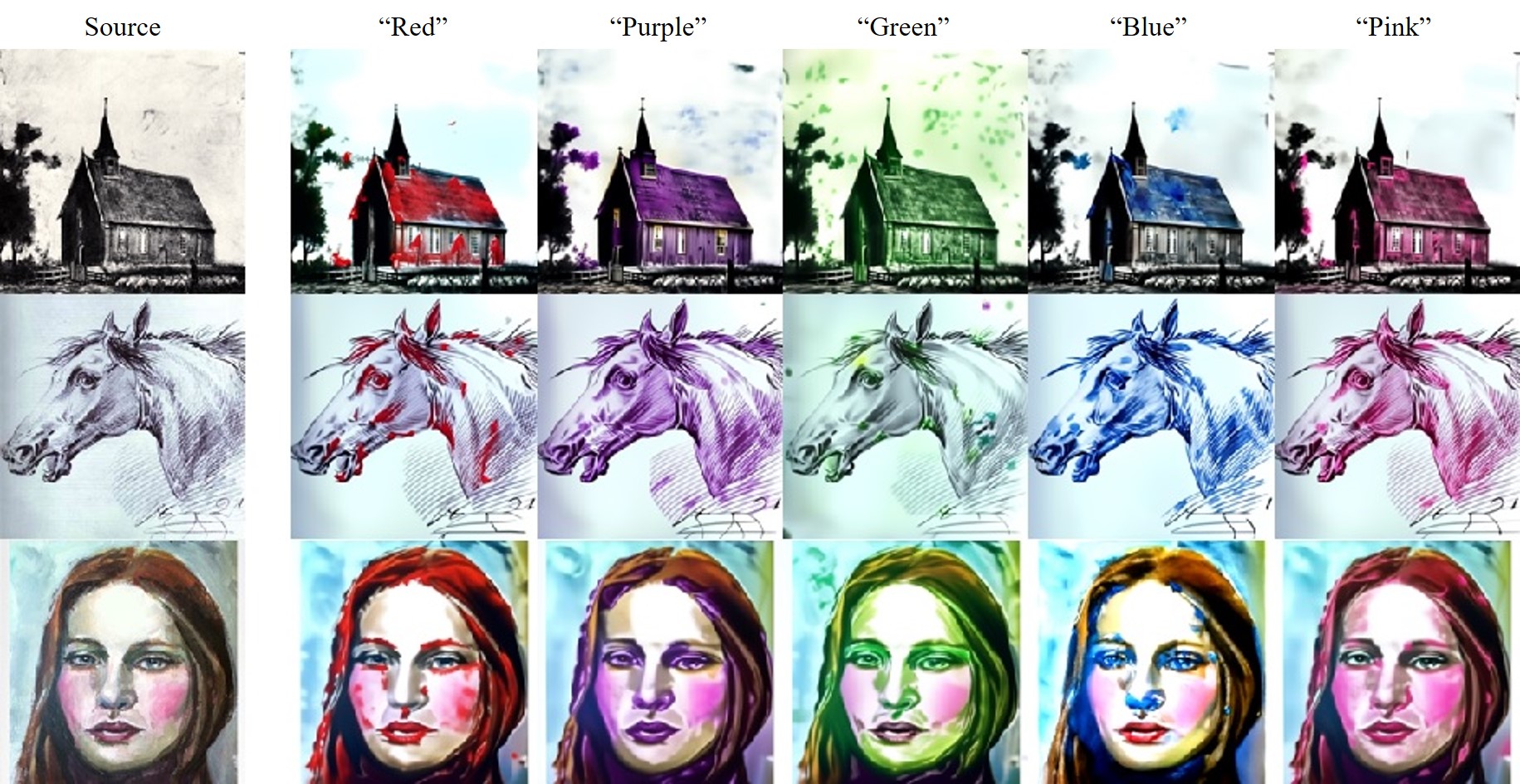}}
	\caption{Additional results on color style prompts. }
	\label{fig_supple_color}
\end{figure*}

\begin{figure*}[!t]
	\centerline{\includegraphics[width=0.8\textwidth]{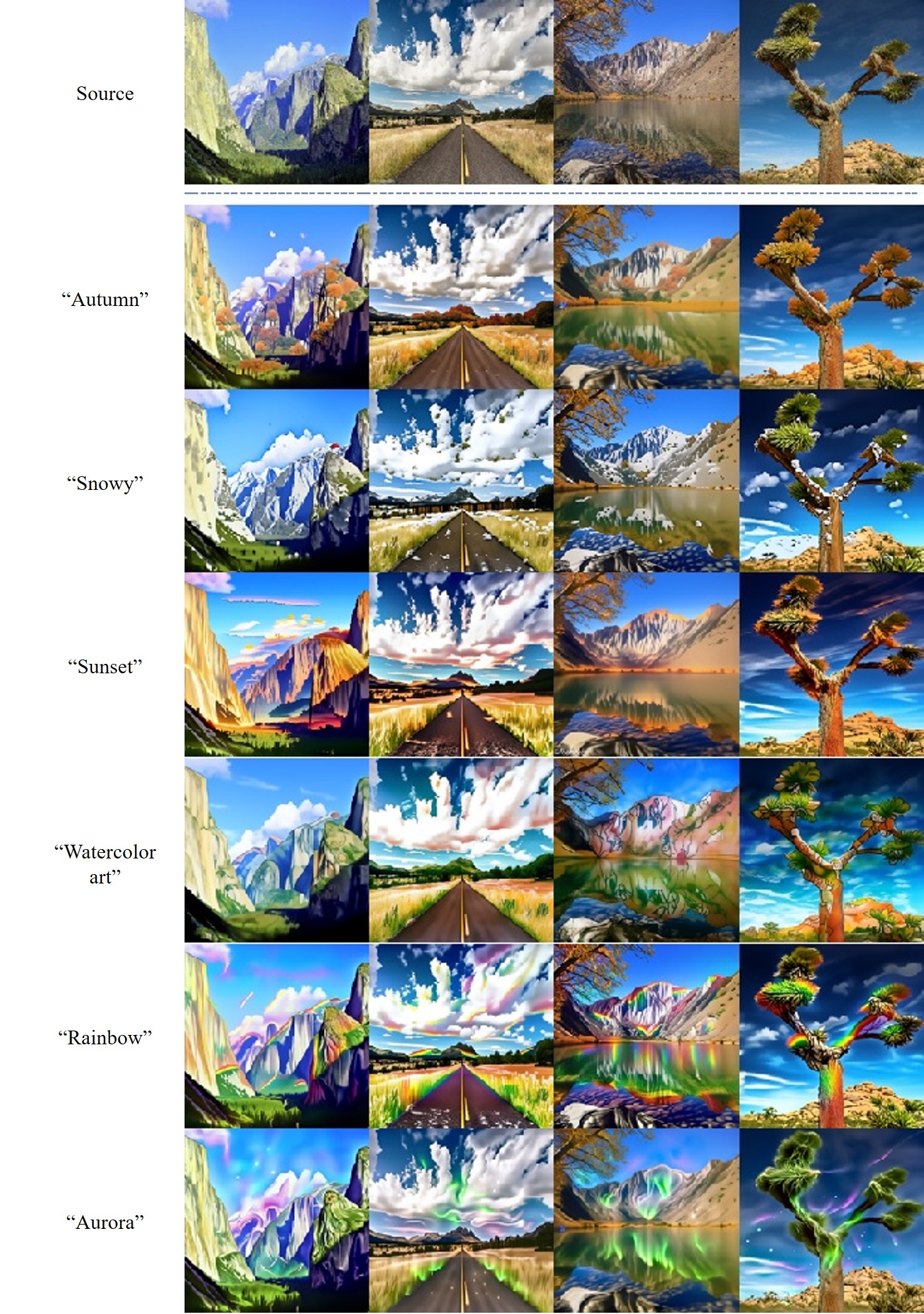}}
	\caption{Additional results on various style prompts. }
	\label{fig_supple_weather}
\end{figure*}

\begin{figure*}[!t]
	\centerline{\includegraphics[width=0.64\textwidth]{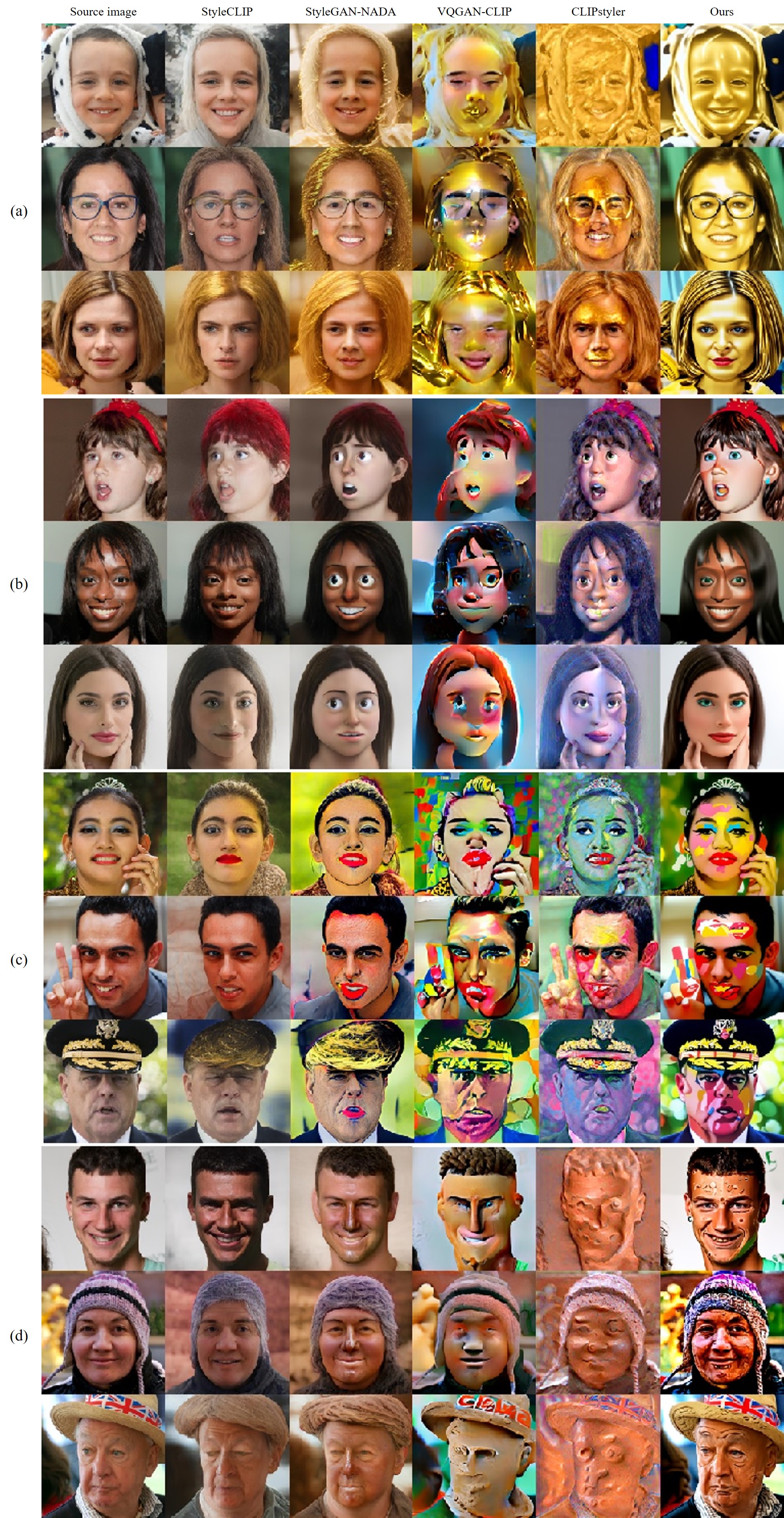}}
	\caption{Additional results on the comparative studies. (a), (b), (c), and (d) are the results on styles of ``golden", ``3d render in the style of Pixar", ``pop art", and ``clay", respectively. }
	\label{fig_supple_gan}
\end{figure*}

\begin{figure*}[!t]
	\centerline{\includegraphics[width=\textwidth]{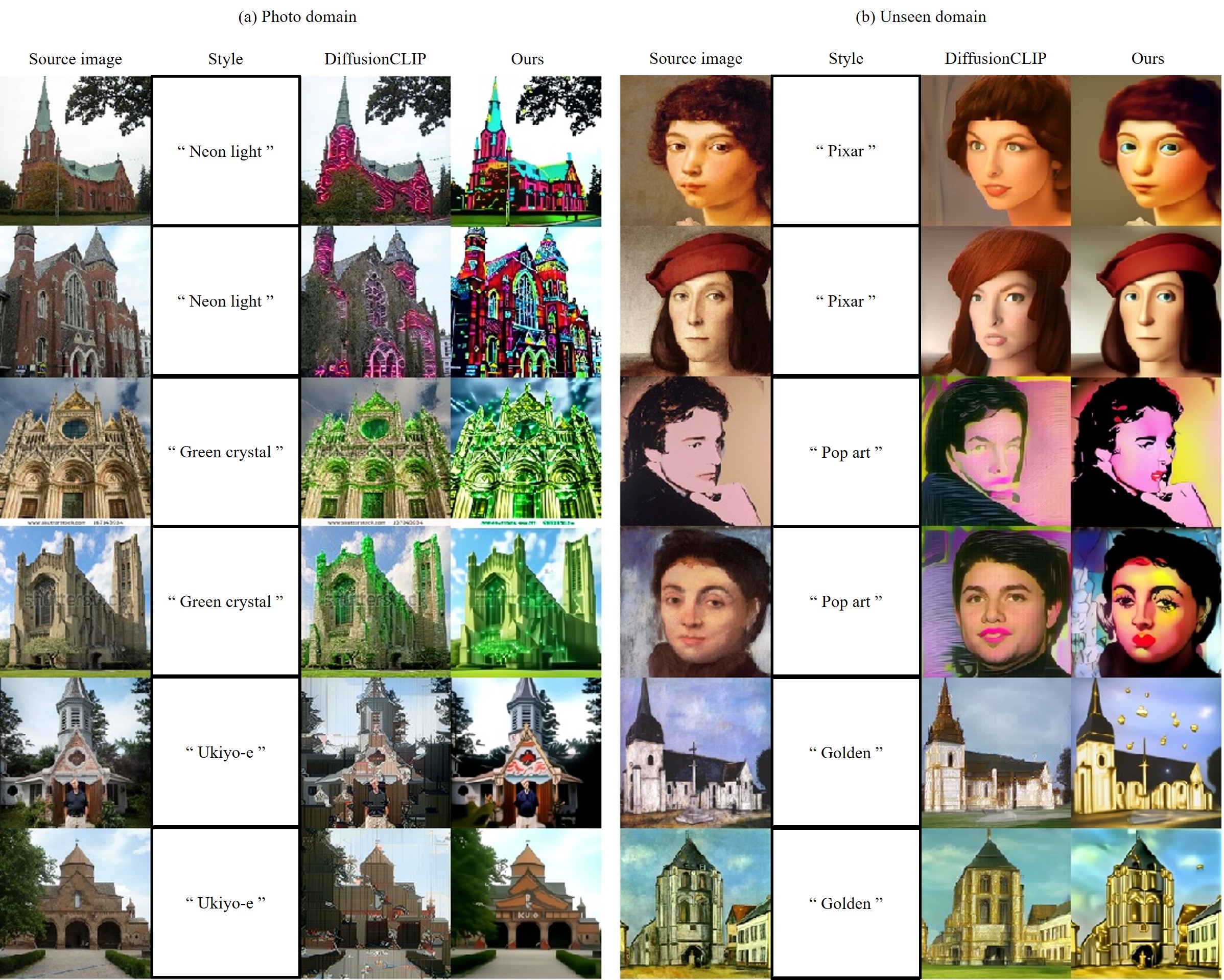}}
	\caption{Additional results on the comparative studies. Source images in (a) are from photo domain and ones in (b) are from unseen domains such as portrait or painting.}
	\label{fig_supple_diffusionclip}
\end{figure*}

\end{document}